\documentclass[lettersize,journal]{IEEEtran}
\usepackage{amsmath,amsfonts}
\usepackage{algorithmic}
\usepackage{algorithm}
\usepackage{array}
\usepackage{bm}
\usepackage{bbm}
\usepackage[caption=false,font=normalsize,labelfont=sf,textfont=sf]{subfig}
\usepackage{textcomp}
\usepackage{stfloats}
\usepackage{url}
\usepackage{verbatim}
\usepackage{graphicx}

\usepackage{xspace}
\usepackage{cite}

\usepackage{array}
\usepackage{booktabs} 
\usepackage{xcolor}
\usepackage{colortbl}

\usepackage{cleveref}
\usepackage{multirow}
\usepackage{booktabs} 
\makeatletter
\DeclareRobustCommand\onedot{\futurelet\@let@token\@onedot}
\def\@onedot{\ifx\@let@token.\else.\null\fi\xspace}
\def\eg{\emph{e.g}\onedot} 
\def\ie{\emph{i.e}\onedot} 

\makeatother
\usepackage{arydshln}
\begin{document}

\title{Negation-Aware Test-Time Adaptation for Vision-Language Models}

\author{Haochen~Han,
        Alex~Jinpeng~Wang,
        Fangming~Liu,
        Jun~Zhu,~\IEEEmembership{Fellow,~IEEE
}
\thanks{
Co-corresponding authors: Alex~Jinpeng~Wang and Fangming~Liu.

Haochen~Han, Alex~Jinpeng~Wang, and Fangming~Liu are with the Department of AI Computing, Pengcheng Laboratory, Shenzhen, China. {e-mail: hhc2077@outlook.com; jinpengwang@csu.edu.cn; fangminghk@gmail.com}. Alex~Jinpeng~Wang is also with the School of Computer Science and Technology, Central South University, Changsha, China.

Jun~Zhu is with the Department of Computer Science and Technology, Institute for AI, BNRist Center, THBI Lab, Tsinghua-Bosch Joint Center for ML, Tsinghua University, Beijing, China. {e-mail: dcszj@tsinghua.edu.cn}.
}}

\markboth{Journal of \LaTeX\ Class Files,~Vol.~14, No.~8, August~2021}%
{Shell \MakeLowercase{\textit{et al.}}: A Sample Article Using IEEEtran.cls for IEEE Journals}



\maketitle

\begin{abstract}
In this paper, we study a practical but less-touched problem in Vision-Language Models (VLMs), \ie, negation understanding. Specifically, many real-world applications require models to explicitly identify what is false or non-existent, \eg, radiologists may search for images that exclude specific conditions. Despite the impressive transferability of VLMs through large-scale training, they suffer from a critical limitation that fails to handle negation. To address this challenge, existing methods attribute its root cause to the scarcity of negation training data and propose to fine-tune VLMs on massive data containing explicit negation. Undoubtedly, such data-centric solutions demand substantial data and computational resources, limiting their sustainable widespread adoption. To tackle negation in a low-carbon manner, we empirically observe that the key obstacle lies in the dual-concept shifts between the affirmation and negation distributions. Therefore, we propose a Negation-Aware Test-Time Adaptation (NEAT) method to efficiently adjust distribution-related parameters during inference. In brief, NEAT can reduce distribution shift in consistent semantics while eliminating false distributional consistency in unrelated semantics. Extensive experiments on the various negation understanding tasks verify the effectiveness of the proposed method. Remarkably, with less than 0.01\% of trainable parameters, NEAT achieves comparable or superior performance to state-of-the-art post-training approaches. Our code is available at \url{https://github.com/hhc1997/NEAT}.

\end{abstract}


\section{Introduction}


Negation, a fundamental logical concept, benefits human perception and decision-making by encoding information about non-existent entities. Recent studies in cognitive science \cite{hasson2006does, de2021look, szabo2022infants} show that humans first understand negation before developing broader world knowledge: even 18-month-olds can use negative sentences to constrain novel object meanings \cite{de2021look}.

However, such cognitive processes in humans do not emerge in advanced multi-modal artificial intelligence, especially Vision-Language Models (VLMs). Conversely, despite learning rich open-world concepts from millions of image-text pairs, VLMs fail to comprehend negation \cite{alhamoud2025vision, park2025know}, severely hindering their applications in many real-world scenarios. For example, drones might query ``a road without ice'' during extreme weather rescue missions, or a radiologist may search for images showing ``pulmonary nodules without malignant features''. Therefore, understanding what is false or non-existent is crucial for VLMs to perform precisely.

To achieve reliability against negation, existing methods \cite{singh2024learn, alhamoud2025vision, park2025know} propose that the root cause lies in the scarcity of negation terms in VLMs' pre-training data, and resort to generating negation-inclusive data for post-training. Specifically, NegBench \cite{alhamoud2025vision} introduces the large-scale synthetic dataset NegFull, which generates more than 70 million image-text pairs containing explicit negation based on CC12M \cite{changpinyo2021conceptual}. The concurrent work NegationCLIP \cite{park2025know} proposes two data generation pipelines using Large Language Model (LLM) and Multi-modal Large Language Model (MLLM) to augment captions with negation. Despite the success, these data-centric approaches require substantial data and computational resources, limiting their sustainable widespread adoption. Thus, it is essential to endow VLMs with negation-aware capabilities in a cost-effective manner.

\begin{figure}[t]
\centering  
\includegraphics[width=\columnwidth]{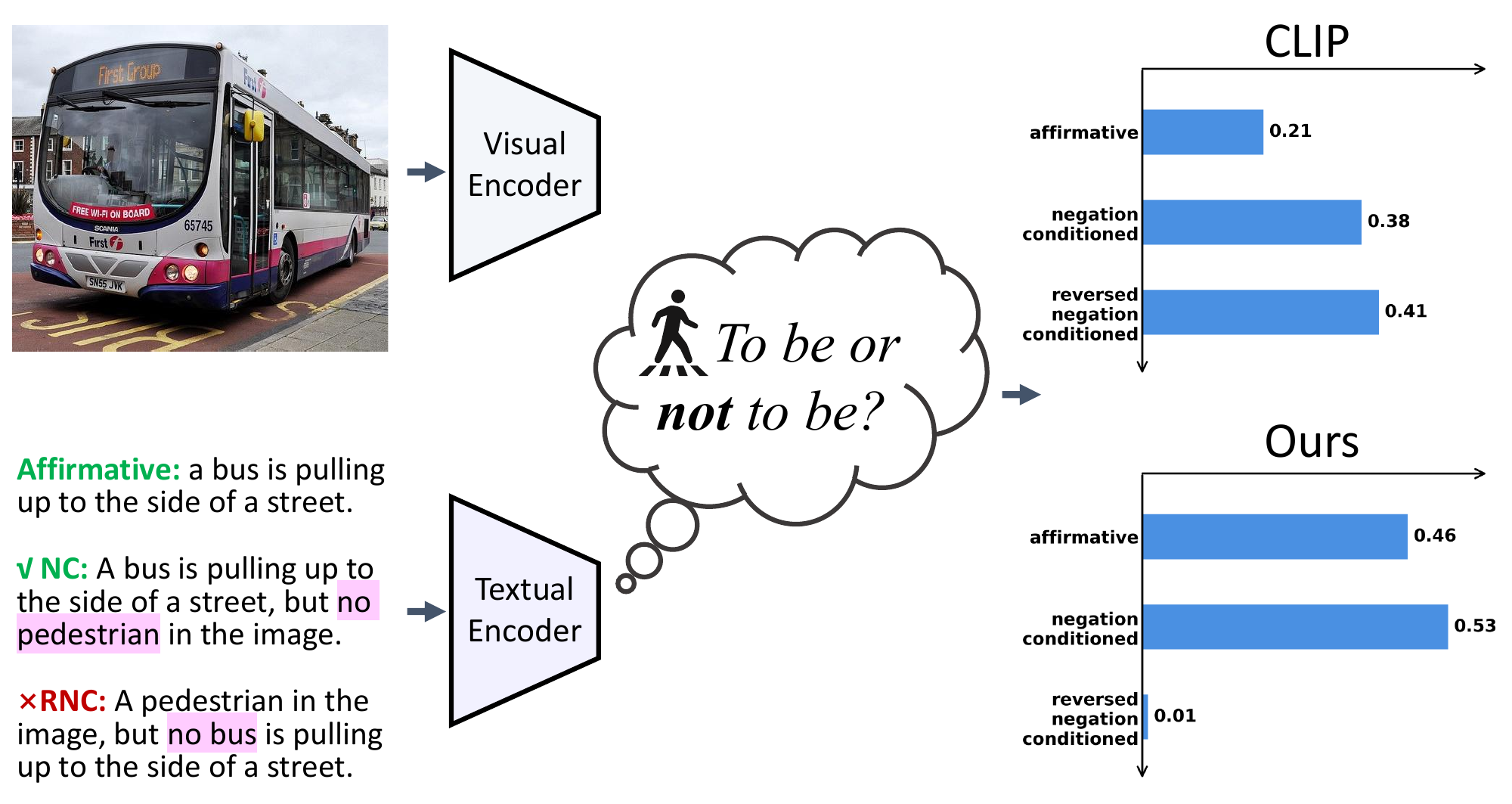}
\caption{A toy example to illustrate the \textbf{dual-concept shifts problem} in Vision-Language Models when understanding negation. Negation-conditioned (NC) texts negate absent elements and share consistent semantics with affirmative ones. Reversed negation-conditioned (RNC) texts wrongly affirm absent elements and negate present ones, which are semantically opposite to NC texts. Although CLIP learns diverse open-world knowledge, it exhibits a similarity gap between affirmative and NC texts, and high similarity between NC and RNC texts. We show how distributional adaptation can tackle this issue.}
\label{fig: intro}
\end{figure}

In this paper, we think outside the box of complex post-training and argue that---\textit{the key obstacle to negation understanding is the dual-concept shifts between the affirmation and negation distributions}. Taking the prevailing CLIP \cite{radford2021learning} for example, as illustrated in Fig. \ref{fig: intro}, CLIP exhibits a significant similarity gap between the affirmative caption and its negation-conditioned caption, \eg, specifying `no pedestrian' for a more precise description, despite them sharing consistent semantics. On the other hand, CLIP shows incorrect similarity between the negation-conditioned caption and its semantically reversed counterpart, \eg, negating existent entities (bus) and wrongly affirming absent ones (pedestrian). As a result, VLMs only interpret negation statements as bags of words and may even produce entirely incorrect judgments.

Based on the above discussions and observations, instead of the full-parameter fitting on negation data, we propose only adjusting distribution-related parameters (\eg, normalization layers) to tackle the dual-concept shifts during inference. To be specific, we propose a novel \textbf{Ne}gation-\textbf{A}ware \textbf{T}est-time Adaptation (dubbed NEAT) framework, which can rapidly learn from unlabeled multi-modal negation data and then make reliable predictions. Our NEAT encapsulates three key components. First, the negation-separated refinement module is adopted to obtain refined entropy between negated captions and visual content, where descriptions are decomposed into two affirmative parts for VLMs to handle adeptly. Second, to address the false distributional consistency in unrelated semantics, we propose to view the reversed negation-conditioned caption as the hardest negative information that shares opposite semantics to its corresponding visual sample. Theoretically, we show that this reversed contrastive objective only requires comparing against limited visual samples and is equivalent to a simple metric loss. Third, to achieve efficient adaptation, we suggest directly reducing the textual-distribution gap arising from dual-concept shifts. Empirically, NEAT establishes state-of-the-art performance on multiple negation benchmarks that span images, videos, and medical scenarios, outperforming the expensive post-trained baselines by a significant margin and observing strong generalization ability in adapted VLMs.

Our main contributions are summarized as follows:

\begin{itemize}
\item To the best of our knowledge, this work could be the first study to enhance negation understanding of VLMs via test-time adaptation. We reveal that the key to negation understanding is the dual-concept shifts problem between affirmation and negation distributions, which could be addressed by adapting lightweight normalization layers of models.

\item We propose a novel test-time adaptation method named NEAT, which comprises three key components: the negation-separated entropy refinement to reduce distribution shift in consistent semantics, the reversed contrastive learning to eliminate the false distributional consistency in unrelated semantics, and the debiased textual learning to further achieve efficient adaptation. 

\item Extensive experiments verify the effectiveness of the proposed method. Remarkably, our NEAT achieves comparable or even superior performance to the best CLIP-based post-training baseline with significantly reduced resources---using less than {0.36\textperthousand} of the data (unlabeled) and {0.014\textperthousand} of the trainable parameters. Moreover, the adapted layers show strong transferability that can be directly generalized to unseen datasets and tasks.
\end{itemize}

\section{Related Works}
\subsection{Negation Understanding in Vision-Language Models} 

Vision-language foundation models learned diverse knowledge from large-scale multi-modal data, which have attracted significant attention due to their powerful transfer capabilities \cite{zhang2024vision, awais2025foundation}. By supervising visual representations with natural language descriptions, pioneering works CLIP \cite{radford2021learning} and ALIGN \cite{jia2021scaling} show great success in various downstream tasks. However, the web-crawled training data are mainly affirmative, which makes it difficult to comprehend the negation of the foundation VLMs. To comprehensively evaluate model performance in negation scenarios, recent works CREPE \cite{ma2023crepe} and CC-Neg \cite{singh2024learn} proposed vision-language benchmarks that target compositional understanding with negation, but the reliance on linguistic templates limits the diversity of real negation queries. In comparison, NegBench \cite{alhamoud2025vision} utilizes LLMs to generate more natural negated captions that span images, videos, and medical datasets. Despite being trained on billion-scale data, the unsatisfactory performance on these benchmarks reveals that VLMs struggle to understand negation. Recent work \cite{zhu2025calling} shows that such problems persist even in powerful MLLMs like GPT-4o and Claude 3.5, which suffer from hallucinations when processing user-provided negation arguments.

To address this limitation, a series of data-centric approaches attempted to introduce negated captions as distractors in an additional post-training phase. For instance, NegCLIP \cite{yuksekgonul2022and} creates targeted negative captions to enhance compositional reasoning. However, these negative descriptions only focus on swapped relations and ignore negation attributes. To this end, CoNCLIP \cite{singh2024learn}  modifies the contrastive loss to incorporate template-based examples that include negation words, \eg, `no', `not', and `without'. The recent advance NegBench \cite{alhamoud2025vision} further fine-tunes VLMs on 70 million synthesized image-text pairs containing diverse negated statements. Concurrent work NegationCLIP \cite{park2025know} proposes two data generation pipelines that leverage LLMs and MLLMs to generate 229k negation-augmented image-text pairs for post-training.

Although these methods have achieved great success, almost all of them assume that the bottleneck of negation understanding lies in the scarcity of training examples with explicit negation. In contrast, this study pioneers an efficient solution from the perspective of the distribution shift. Considering the computational burden of large VLMs, our method can directly adapt VLMs to handle various negation understanding tasks with only a few parameter updates at test time.

\subsection{Test-Time Adaptation}
The distribution shift between training and test data poses a key challenge when transferring the zero-shot capabilities of foundation VLMs \cite{wang2023enhancing, guo2024tuning, luo2024moil}. In practice, such shifts are inevitable due to environmental variations \cite{koh2021wilds} or the encounter of unseen concepts \cite{xiao2024any}. To address the problem, considerable efforts have been devoted to developing training-time solutions: domain adaptation methods \cite{hao2023dual} that narrow gaps between source and target samples during training. Domain generalization methods \cite{addepalli2024leveraging} that directly learn domain-invariant representations through robust training strategies. However, these methods may require access to source domain data and cannot achieve adaptation in an online manner,  limiting their practicality in real-time applications such as autonomous driving.

To this end, test-time adaptation methods have emerged to adapt pre-trained VLMs to test samples on the fly. The most prevalent application is to tackle out-of-distribution images \cite{karmanov2024efficient, shu2022test, feng2023diverse, yoon2024c, zhao2024test, lee2025ra, zhang2024dual} caused by corruptions or environments. For example, TDA \cite{karmanov2024efficient} designs a training-free dynamic adapter to gradually refine pseudo labels of test samples. TPT \cite{shu2022test} learns adaptive prompts by enforcing entropy consistency across augmented views with confidence-based selection. DiffTPT \cite{feng2023diverse} leverages pre-trained diffusion models to generate diverse and informative data augmentations to improve TPT. C-TPT \cite{yoon2024c} reduces the prediction uncertainty of TPT by establishing calibration error with text feature dispersion. RLCF \cite{zhao2024test} introduces a CLIP model to provide feedback that avoids the pitfall of the entropy minimization. RA-TTA \cite{lee2025ra} uses external knowledge obtained from a web-scale image database to prevent over-reliance on model predictions. DPE \cite{zhang2024dual} evolves prototypes from both textual and visual modalities to capture more accurate multi-modal representations. Beyond these zero-shot image classification tasks, recent works have also shown the promise of TTA in more challenging vision-language retrieval tasks. For instance, to alleviate the potential bias from gender or race, PBM \cite{kong2024mitigating} generates fair retrieval subsets from the self-prediction of VLMs. TCR \cite{li2024test} manipulates both the modality uniformity and modality gap to achieve robust retrieval under query shift.

Different from these prior arts that mainly focus on covariate shifts like corruptions (\eg, noise, blur, and weather) or style variations (\eg, cartoon and sketch), this paper presents the first attempt to tackle concept shift problems arising from negation understanding. As negation can enable more precise queries in many real-world scenarios, it is reasonable to believe that this study could provide some novel and practical insights to the multi-modal community.

\begin{figure*}[htbp]
\centering  
\includegraphics[width=0.95\textwidth]{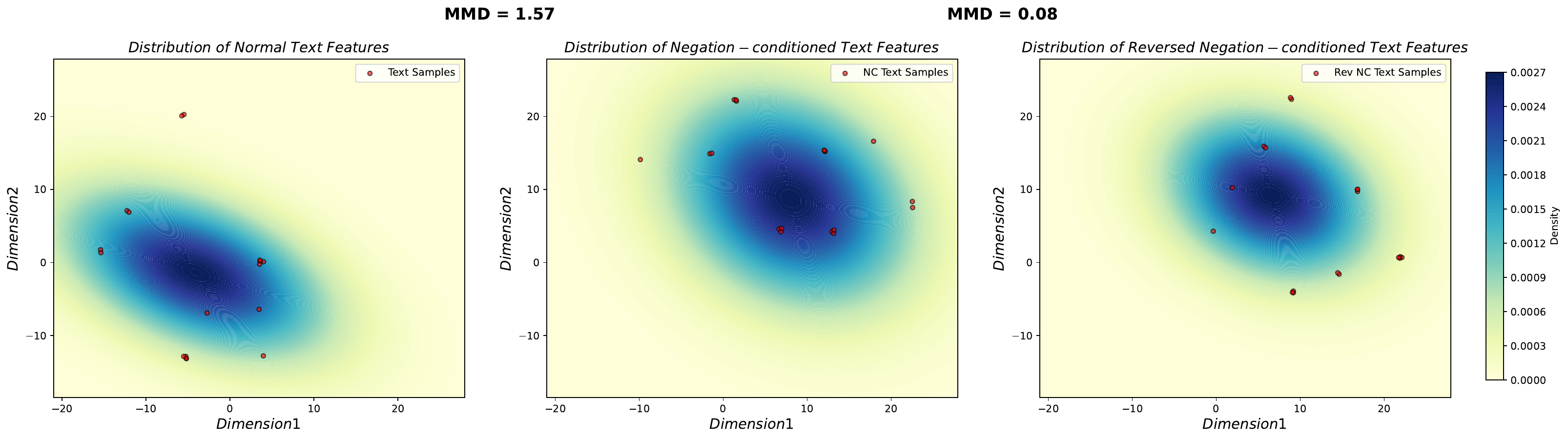}
\caption{\textbf{Dual-concept shifts} problem observed in pre-trained VLMs when understanding negation. Distributions of normal text, negation-conditioned text, and reversed negation-conditioned text of the test set of CIFAR10 are shown on the left, middle, and right, respectively. All embeddings are produced by the OpenAI CLIP and visualized with UMAP \cite{mcinnes2018umap} for dimension reduction. The Maximum Mean Discrepancy (MMD) metrics are also provided above for quantitative analysis. The distributions of normal and negation-conditioned text show a strong shift (MMD=$1.57$) despite sharing the same semantics. Meanwhile, the distributions of negation-conditioned and its reversed text show high consistency (MMD=$0.08$) despite being semantically opposite.}
\label{fig: key_observation}
\end{figure*}

\section{Preliminaries}
\subsection{VLMs Do Not Understand Negation}
Let $f_\theta$ represent a VLM pre-trained on image-text pairs $\mathcal{D}_{train} = \{\mathcal{V}, \mathcal{T}\}$, where $v_i \in \mathcal{V}$ is the raw image and $t_i \in \mathcal{T}$ is the corresponding textual caption. One fundamental capability of VLMs is to make multiple data modalities comparable in the same embedding space, where visual embedding $\bm{v}_i \in \mathbb{R}^D$ and textual embedding $\bm{t}_i \in \mathbb{R}^D$ are derived by passing $v_i$ and $t_i$ through its visual encoder $f_{\theta_{V}}$ and textual encoder $f_{\theta_{T}}$, respectively.

Despite the promising performance through large-scale training, the textual supervision $\mathcal{T}$ in data is mainly expressed affirmatively, which poses a natural limitation: vision-language foundation models fail to understand negation.

\subsection{Test-Time Adaptation for Negation Understanding}
For the pre-trained VLM, the key obstacle to negation understanding is the incongruence between the train and test distributions.  Specifically, the text with negation conditions can be viewed as a concept shift \cite{xiao2024any} compared to the affirmative expressions, \ie, $P(\mathcal{T}_{test}) \neq P(\mathcal{T})$, where $P(\cdot)$ denotes the distribution of the given textual data. As a result, VLMs suffer performance degradation \cite{alhamoud2025vision} when facing real-world applications requiring comprehension of negation. To address such a generalization problem, TTA has emerged to boost the foundation model under data distribution shifts, which online updates only a minimal set of $\theta$, \eg, normalization layers \cite{wang2021tent} or prompt vectors \cite{shu2022test}, at test time before making a prediction.

\section{Methodology}
\subsection{Analysis of Concept Shifts in Negation Understanding}\label{subsec:concept-shifts}
In this section, we present the analysis of where the shift in negation understanding comes from. Let $\hat{t}_i$ denote a text constrained by some negation conditions that maintains semantic consistency with its affirmative counterpart $t_i$. For instance, $t_i$ typically describes the object in $v_i$ or its associated attributes, \eg, ``a photo of a dog''. By introducing some negative concepts, $\hat{t}_i$ can exclude specific elements for a more fine-grained description, \eg, ``a photo of a dog not on grass'' or "a photo of a dog not running". To explore how such negation affects the understanding of VLMs, we experiment with the representative CLIP model on CIFAR10 and have several nontrivial findings.

\textbf{Distribution shift within semantic consistency.} For each image, we first form the normal caption ``a photo of the [CLASS]'' by inserting its class name into a fixed prompt template. Then, we synthesize a corresponding negation-conditioned caption ``a photo of the [CLASS] but not of the [CLASS$^{\prime}$]'' by randomly incorporating a different class name. As each image in CIFAR10 is single-categorized, both descriptions are correct and ideally share consistent semantics. However, as shown in Fig. \ref{fig: key_observation}, the distributions of normal text (Fig. \ref{fig: key_observation} left) and negation-conditioned text (Fig. \ref{fig: key_observation} middle) show a significant shift in the embedding space.

\textbf{Distribution consistency within semantic shift.} For each negation-conditioned caption, we synthesize a semantically-reversed counterpart ``a photo of the [CLASS$^{\prime}$] but not of the [CLASS]'' by simply swapping two class names. Obviously, such two descriptions convey entirely opposite semantics. However, we observed that the distributions of negation-conditioned text and their semantically reversed text (Fig. \ref{fig: key_observation} right) show a substantial similarity in the embedding space.

Despite containing rich and diverse knowledge, these observations indicate that pre-trained VLMs only interpret text as bags of words \cite{yuksekgonul2022and}. Consequently, VLMs behave with the above dual-concept shifts when understanding negation. 

\begin{figure*}[t]
\centering  
\includegraphics[width=0.92\textwidth]{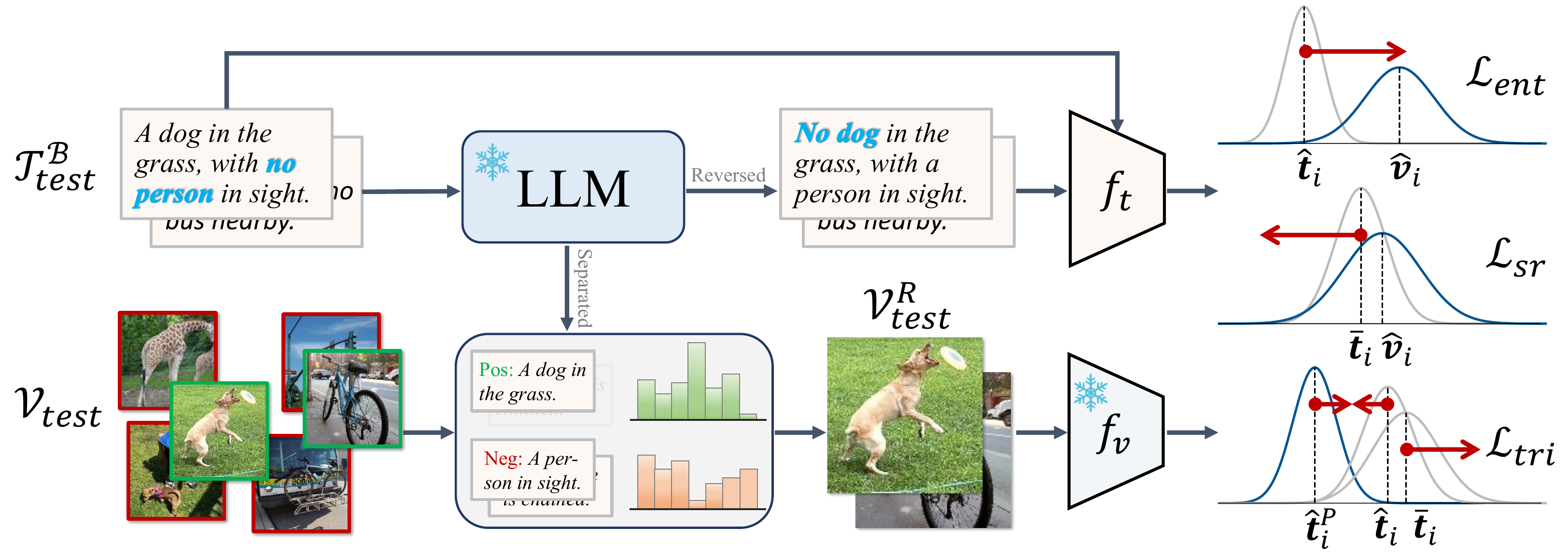}
\caption{Overview of the proposed NEAT. Given the test data with negation contexts, a LLM is first employed to separate the negation forms and generate semantically reversed counterparts. Then the refinement module is used to coarsely select candidate visual samples that are similar to positive entities while distant from negative entities. After that, three training objectives are adopted to reduce distribution shift in consistent visual-textual semantics ($\mathcal{L}_{ent}$), eliminate the false distributional consistency in unrelated visual-textual semantics ($\mathcal{L}_{sr}$), and directly debias the dual-concept shifts in the textual modality ($\mathcal{L}_{tri}$).}
\label{fig: framework}
\end{figure*}

\subsection{Negation-Aware Test-Time Adaptation}\label{subsec: Refined P}
Next we describe our \textbf{Ne}gation-\textbf{A}ware \textbf{T}est-time Adaptation (NEAT) framework in detail. As shown in Fig. \ref{fig: framework}, NEAT comprises three key components to bridge the above dual-concept shifts, \ie, the refined entropy minimizing loss $\mathcal{L}_{ent}$, the semantics reversion loss $\mathcal{L}_{sr}$, and the textual debiasing loss $\mathcal{L}_{tri}$. For the given online batch with size $\mathcal{B}$, we update the pre-trained VLM in time by the following training objective:
\begin{equation}\label{eq: NEAT objective}
\min_{\tilde{\theta}_{T}}\mathcal{L}(\mathcal{B}) = \mathcal{L}_{ent} + \mathcal{L}_{sr} + \mathcal{L}_{tri},
\end{equation}
where $\tilde{\theta}_{T} \subseteq \theta_{T}$ denotes the parameters of normalization layers to only adjust the data distribution.  In the following, we will elaborate on each loss individually.

\subsubsection{Entropy Refinement via Negation Separation}\label{subsec:refined-entropy}
In online TTA, we have an unlabeled vision-language dataset $\mathcal{D}_{test} = \{\mathcal{V}_{test} = \{v_i \}_{i=1}^{N_V}, \mathcal{T}_{test} = \{\hat{t}_i \}_{i=1}^{N_T}\}$, where $v_i$ depicts the natural visual concepts like in $\mathcal{D}_{train}$ but $\hat{t}_i$ may contain additional negative conditions that differ from $\mathcal{D}_{train}$. Our goal is to adapt the pre-trained model immediately to establish semantic correspondence between $\mathcal{V}_{test}$ and $\mathcal{T}_{test}$. To achieve this, one simple yet powerful solution is to fine-tune $f_{\theta_{T}}$ by minimizing the entropy of its predictions \cite{wang2021tent}. Formally, denote $\bm{T}_{test}^{\mathcal{B}} = [\hat{\bm{t}}_1, \dots, \hat{\bm{t}}_{\mathcal{B}}]^\top \in \mathbb{R}^{{\mathcal{B}} \times D}$ the batched textual features generated by $f_{\theta_{T}}$, and $\bm{V}_{test} = [\bm{v}_1, \dots, \bm{v}_{N_T}]^\top \in \mathbb{R}^{{N_T} \times D}$ the fixed visual features generated by $f_{\theta_{V}}$, the adaptation object of the online batch is
\begin{equation}\label{eq: entropy}
\min_{\tilde{\theta}_{T}}\mathcal{L}_{ent} = -\frac{1}{N_T} \sum_{i=1}^{N_T} \bm{P}_i \log \bm{P}_i, 
\end{equation}
where $\bm{P}_i = \text{Softmax}\left(\bm{v}_1\left(\bm{T}_{test}^{\mathcal{B}}\right)^\top / \tau_1 \right) \in \mathbb{R}^{\mathcal{B}}$ is the model’s output probability smoothed by a temperature $\tau_1$. However, the large computational size would make entropy a sub-optimal confidence metric, leading to either model underfitting or overfitting \cite{li2024test}.

To alleviate this, we follow the candidate selection strategy suggested in \cite{li2024test} to refine the prediction of the model. Specifically, let $\mathcal{N}(\cdot)$ denote a selection manner, and the corresponding visual candidate in the batch is obtained by
\begin{equation}\label{eq: visual candidates}
\bm{V}_{test}^{R} = [\hat{\bm{v}}_1, \dots, \hat{\bm{v}}_{\mathcal{B}}]^\top \in \mathbb{R}^{\mathcal{B} \times D}, \hat{\bm{v}}_i = \mathcal{N}(\hat{\bm{t}}_i), \forall i \in [1,\mathcal{B}].
\end{equation}
The most straightforward selection method is the nearest neighborhood based on similarity \cite{li2024test}. However, as mentioned in Section \ref{subsec:concept-shifts}, the strong shift within the training and test patterns makes the spatial relation of embeddings unreliable.

As a remedy, we propose to separate the negation part from $\hat{t}_i$ to avoid model understanding of negative semantics. To this end, we utilize the in-context learning ability of LLMs to decompose $\hat{t}_i$ into positive component $\hat{t}_i^P$ and negative one $\hat{t}_i^N$. For example, ``a photo of a dog not on grass'' would be split into ``a photo of a dog'' and ``a photo of grass'', where the negative element is also recaptioned affirmatively. Such simple parse tasks could be handled by lightweight LLMs, and we empirically find that Llama-3-8B \cite{dubey2024llama} performs well for our goal. As the pre-trained VLM can readily match visual samples to affirmative captions, we can use the similarity between $v_j$ and $\hat{t}_i^N$ to penalize the correspondence from $v_j$ to $\hat{t}_i^P$:
\begin{equation}\label{eq: decomposed sim}
S(\hat{\bm{t}}_i, \bm{v}_j) = (\hat{\bm{t}}_i^P \bm{v}_j^{\top})(1-\alpha[\hat{\bm{t}}_i^N \bm{v}_j^{\top}]_+),
\end{equation}
where $\alpha$ is a trade-off parameter to control the penalty and $[x]_+ = max(x,0)$ is the hinge function. Intuitively, higher $S(\hat{\bm{t}}_i, \bm{v}_j)$ requires ${v}_j$ depicts $\hat{t}_i^P$ well and remains unrelated to concepts in $\hat{t}_i^N$. Then we select the most similar sample by Eq.\eqref{eq: decomposed sim} to obtain the corresponding visual candidates $\bm{V}_{test}^{R}$. Consequently, the entropy minimizing objective for the online batch is
\begin{equation}\label{eq: refined prediction}
\begin{aligned}
\min_{\tilde{\theta}_{T}}\mathcal{L}_{ent} &= -\frac{1}{\mathcal{B}} \sum_{i=1}^{\mathcal{B}} \bm{P}_i^R \log \bm{P}_i^R, 
\\ \text{where} \ \bm{P}_i^R& = \text{Softmax}\left(\hat{\bm{v}}_1\left(\bm{T}_{test}^{\mathcal{B}}\right)^\top / \tau_1 \right) \in \mathbb{R}^{\mathcal{B}}.
\end{aligned}
\end{equation}
By excluding irrelevant samples, the refined entropy could prevent model underfitting and enhance negation understanding through narrowing undesired concept bias.

\subsubsection{Semantics Reversion Learning}\label{subsubsec:sr learning}
Section \ref{subsec:refined-entropy} presents a solution to reduce distribution shift in consistent semantics, yet the incorrect distribution consistency still exists in unrelated semantics. To overcome this challenge, we propose to discriminate these entangled distributions using the hardest negation. Specifically, for each $\hat{t}_i$, we first employ the LLM to construct a semantically reversed caption $\bar{t}_i$ by swapping its positive and negative components. For example, ``a photo of a dog not on grass'' would be reversed as ``a photo of grass but not of a dog''. Ideally, on the one hand, $\bar{t}_i$ is completely irrelevant to its visual candidate $\hat{v}_i$ since it describes absent objects while negating present ones. On the other hand, $\bar{t}_i$ maintains certain similarity to the batched-unpaired sample $\hat{v}_j$, since it correctly excludes those absent objects. Such learning objectives  naturally are equivalent to minimizing the mutual information between representations of paired $\bar{t}$ and $\hat{v}$, which could be approximated as the negative InfoNCE \cite{oord2018representation}:
\begin{equation}\label{eq: neg InfoNCE}
\min_{\tilde{\theta}_{T} }\mathbb{E}_{p(\bar{t},\hat{v})} \left[ \bar{\bm{t}} \hat{\bm{v}}^{\top}/\tau_2 - \log \mathbb{E}_{p(\hat{v}^{\prime})} \exp(\bar{\bm{t}} \hat{\bm{v}}^{\prime\top}/\tau_2) )\right],
\end{equation}
where the pair $(\bar{t},\hat{v})$ is sampled from the joint distribution $p(\bar{t},\hat{v})$ and the pair $(\bar{t},\hat{v}^{\prime})$ is built by samples independently sampled from the marginal $p(\hat{v}^{\prime})$, and $\tau_2$ is a temperature parameter.

\begin{figure}[t]
\centering  
\includegraphics[width=\columnwidth]{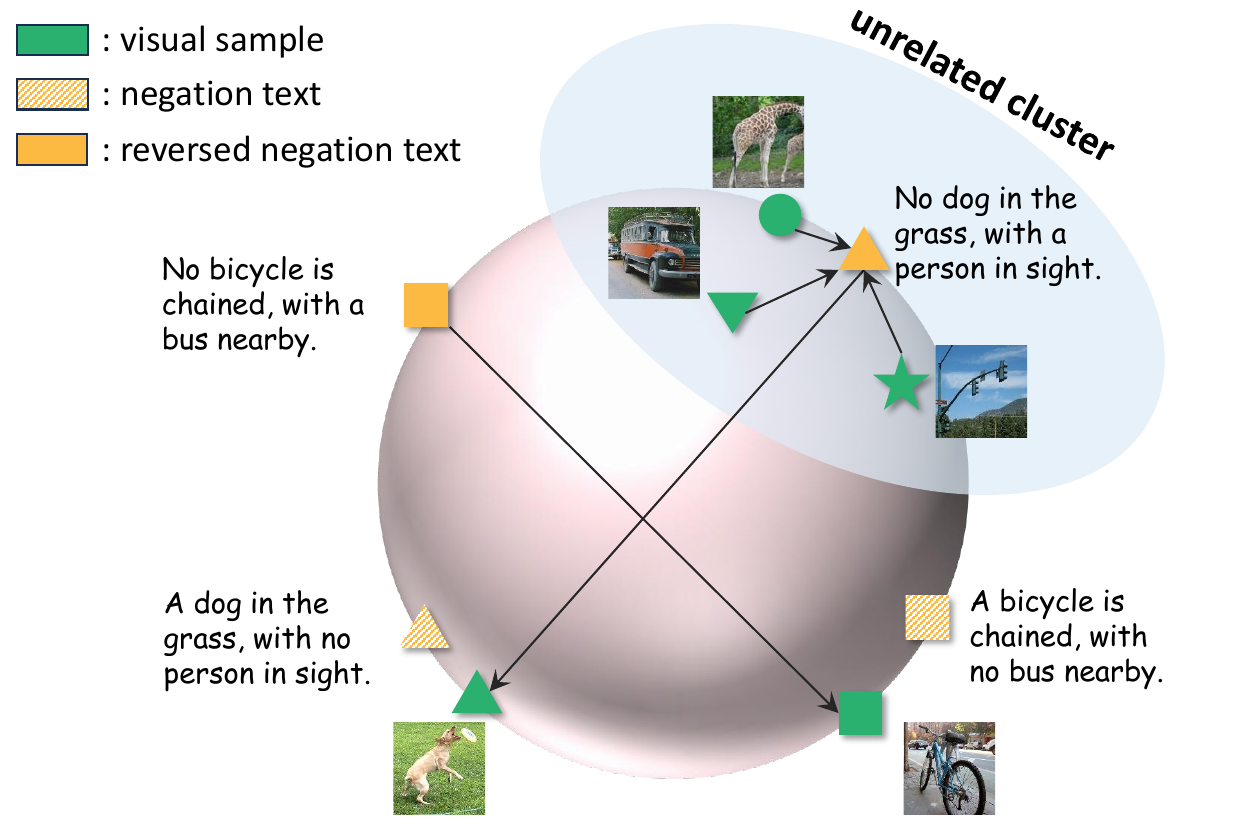}
\caption{A toy example to show the challenge of semantic-reversed contrastive learning. Although the negative InfoNCE pushes the reversed negation-conditioned text away from the target image, it relatively brings unrelated images closer together, thus corrupting the learned semantic structure. }
\label{fig: Neg_InfoNCE}
\end{figure}

As illustrated in Fig. \ref{fig: Neg_InfoNCE}, although minimizing Eq.\eqref{eq: neg InfoNCE} can force the paired samples to be distant, bringing the reversed text closer to all unpaired visual samples would corrupt the learned semantic structure. To prevent the clustering of unrelated visual samples, we propose to sample only the hardest visual sample $\hat{v}_{i}^{-}$ that shares minimal similarity with the reversed text $\bar{t}_i$. As $\bar{t}_i$ provides partial supervision by specifying non-existent concepts, it is reasonable to align  $\bar{t}_i$ with the most irrelevant visual content. Then the semantics reversion objective for the online batch is:
\begin{equation}\label{eq: contrastive object}
\begin{aligned}
\min_{\tilde{\theta}_{T}}&\mathcal{L}_{sr} =\frac{1}{\mathcal{B}} \sum_{i=1}^{\mathcal{B}} \log \left[ \frac{\exp (\bar{\bm{t}}_i \hat{\bm{v}}^{\top}_i/\tau_2)}{\exp (\bar{\bm{t}}_i \hat{\bm{v}}^{\top}_i/\tau_2) + \exp (\bar{\bm{t}}_i \hat{\bm{v}}^{-\top}_i/\tau_2)} \right]
\\ &= \frac{1}{\mathcal{B}} \sum_{i=1}^{\mathcal{B}} \log \left[\frac{1}{1 + \exp \left[({\bar{\bm{t}}_i \hat{\bm{v}}^{-\top}_i - \bar{\bm{t}}_i \hat{\bm{v}}^{\top}_i})/\tau_2 \right]} \right]
\\ &= \frac{1}{\mathcal{B}} \sum_{i=1}^{\mathcal{B}}-\log\left[1 + \exp\left[({\bar{\bm{t}}_i \hat{\bm{v}}^{-\top}_i - \bar{\bm{t}}_i \hat{\bm{v}}^{\top}_i})/\tau_2 \right]\right]
\\ &\simeq \frac{1}{\mathcal{B}} \sum_{i=1}^{\mathcal{B}}[(\bar{\bm{t}}_i \hat{\bm{v}}^{\top}_i - \bar{\bm{t}}_i \hat{\bm{v}}^{-\top}_i)/\tau_2].
\end{aligned}
\end{equation}
Intuitively, Eq.\eqref{eq: contrastive object} forces the reversed text to provide stronger negative supervision to its corresponding visual sample, which can eliminate the undesired distribution consistency.

\subsubsection{Textual Dual-Concept Debiasing}
The above training objectives mitigate the dual-concept shifts in cross-modal alignment. We further strengthen this debiasing in textual modality  by directly regulating the triplet semantic gap:
\begin{equation}\label{eq: textual debiasing}
\min_{\tilde{\theta}_{T}}\mathcal{L}_{tri} = \frac{1}{\mathcal{B}} \sum_{i=1}^{\mathcal{B}} \left[ \lambda\|\hat{\bm{t}}_i - \hat{\bm{t}}_i^{P}\|_2 +(2-\|\hat{\bm{t}}_i - \bar{\bm{t}}_i\|_2) \right],
\end{equation}
where $\lambda$ is the balance hyperparameter. In Eq.\eqref{eq: contrastive object}, the first term brings the negation-conditioned caption closer to its affirmative, while the second term encourages maximum distance between the negation-conditioned caption and its reversed counterpart.

\section{EXPERIMENTS}
In this section, we evaluate the effectiveness of our proposed method through extensive experiments, comparing it with both state-of-the-art post-training and test-time adaptation methods. For a comprehensive comparison, our experiments are conducted across diverse visual domains, including images, videos, and medical imaging.

\begin{table}[h]

  \centering
    \renewcommand{\arraystretch}{1}
      \caption{Summary of datasets and tasks for negation understanding.}
    \resizebox{1.0\columnwidth}{!}{
    \begin{tabular}{llcc}
    \hline
    \hline
    Dataset & Task  & Size  & Negation Type \\
    \midrule
    \multirow{2}[2]{*}{COCO} &       Retrieval-Neg & 5,000  & LLM-Rephrased \\
          & MCQ-Neg & 5,914  & LLM-Rephrased \\
    \midrule
    VOC2007 & MCQ-Neg & 5,032  & LLM-Rephrased \\
    \midrule
    \multirow{2}[2]{*}{MSR-VTT} & Retrieval-Neg & 1000  & LLM-Rephrased \\
          & MCQ-Neg & 1,000  & LLM-Rephrased \\
    \midrule
    \multirow{1}[2]{*}{CheXpert} & Binary Classification-Aff  & 2,352  & Templated \\
          & Binary Classification-Neg & 616   & Templated \\
    \hline
    \hline
    \end{tabular}%
    }

  \label{tab: datasets}%
\end{table}%

\subsection{Experiment Setting}

\subsubsection{Datasets} Following the recent NegBench \cite{alhamoud2025vision} evaluations, we test NEAT on four datasets: COCO, VOC2007, MSR-VTT, and CheXpert, each containing one or more well-designed negation understanding tasks. Specifically, 
\begin{itemize}
\item \textbf{Retrieval-Neg task} evaluates whether VLMs can tackle real-world queries that mix affirmative and negative statements, such as `a street without cars' or 'flowers that are not red'. This task challenges the model to not only identify the present elements but also exclude specific content, which could simulate more fine-grained retrieval in search engines and recommendation systems.

\item \textbf{MCQ-Neg task} requires VLMs to select the correct description of the given image from multiple closely related choices. Based on the linguistic template, alternative descriptions could be categorized as Affirmation, Negation, and Hybrid. This task challenges the model to parse subtle yet critical differences among hard negatives.

\item \textbf{Binary Classification-Neg task} tests whether VLMs can correctly interpret negation words by requiring models to distinguish the presence or absence of specific elements. This task is essential in medical diagnostics. For instance, `The X-ray shows no evidence of pneumonia' is typical for ruling out lung pathologies.
\end{itemize}

\begin{table*}[t]
  \centering
  \setlength{\aboverulesep}{-0.9pt}
\setlength{\belowrulesep}{0pt}
\setlength{\extrarowheight}{0pt}
  \renewcommand\arraystretch{1}
  \addtolength{\tabcolsep}{-1pt}
    \caption{Comparisons with state-of-the-art post-training methods on MS-COCO and MSR-VTT datasets under negated queries. All methods are based on the ViT-B/32 architecture for the image encoder. The best results are marked in \textbf{bold}.}
  \resizebox{1.0\textwidth}{!}{
    \begin{tabular}{l|l|cccccc|l|cccccc|l}
 \toprule
    \multirow{1.4}[6]{*}{\ Method} & \multirow{1.4}[6]{*}{\ \ Fine-tune Data} & \multicolumn{7}{c|}{\textbf{Retrieval-Neg Task}}               & \multicolumn{7}{c}{\textbf{Retrieval Task}} \\
\cline{3-16}          &       & \multicolumn{3}{c|}{Image-to-Text} & \multicolumn{3}{c|}{Text-to-Image} & \multirow{1}[4]{*}{\ \ \ \ rSum} & \multicolumn{3}{c|}{Image-to-Text} & \multicolumn{3}{c|}{Text-to-Image} & \multirow{1}[4]{*}{\ rSum} \\
\cline{3-8}\cline{10-15}          &       & R@1   & R@5   & \multicolumn{1}{c|}{R@10} & R@1   & R@5   & R@10  &       & R@1   & R@5   & \multicolumn{1}{c|}{R@10} & R@1   & R@5   & R@10  &  \\
 \hline
\multicolumn{16}{c}{\textit{Evaluation Dataset: MS-COCO}} \\
 \hline
\multirow{3}[1]{*}{CLIP} & None  & 45.30 & 69.78 & 79.10 & 24.96 & 47.93 & 59.40 & 326.47 & 49.98 & 75.02 & 83.30 & 30.37 & 54.76 & 66.11 & 359.54 \\
      & CC12M$\sim12M$ & 47.66 & 72.60 & 81.28 & 29.00 & 53.71 & 64.72 & 348.97 & \textbf{52.14} & 75.38 & \textbf{83.94} & \textbf{33.87} & \textbf{58.90} & \textbf{69.45} & \textbf{373.68} \\
      & NegFull$\sim70M$ & 44.48 & 69.32 & 78.40 & 28.17 & 51.91 & 62.98 & 335.26 & 48.60 & 72.86 & 81.08 & 30.07 & 54.19 & 65.32 & 352.12 \\

\rowcolor[rgb]{ 1.0,  0.925,  0.905} & None  & 48.98 & \textbf{74.60} & \textbf{82.14} & 30.00 & 54.63 & 65.66 & $356.01{{\uparrow_{29.5}}}$ & 50.74 & \textbf{76.02} & 83.74 & 31.11 & 55.89 & 66.95 & 364.45 \\
\rowcolor[rgb]{ 1.0,  0.925,  0.905} \multirow{-1.8}[0]{*}{+NEAT} & CC12M$\sim12M$ & \textbf{49.04} & 73.70 & 81.70 & \textbf{31.91} & \textbf{56.70} & \textbf{67.72} & $\textbf{360.77}{{\uparrow_{11.8}}}$ & 50.96 & 74.54 & 82.88 & 33.25 & 58.22 & 69.08 & 368.93 \\
       \hdashline[3pt/3pt]
    ConCLIP & CC-Neg$\sim0.23M$ & 43.04 & 66.64 & 76.12 & 24.30 & 48.22 & 59.83 & 318.15 & 45.72 & 68.44 & 76.16 & 27.20 & 51.93 & 63.50 & 332.95 \\
     \hdashline[3pt/3pt]
    \multirow{3}[0]{*}{NegCLIP} & None  & 51.64 & 77.92 & 86.06 & 37.02 & 64.39 & 74.77 & 391.80 & 56.82 & 80.70 & 88.16 & 41.60 & 68.73 & 78.91 & 414.92 \\
          & CC12M$\sim12M$ & 54.44 & 79.72 & 86.94 & 38.35 & 65.38 & 75.66 & 400.49 & 58.68 & 82.56 & 89.38 & 42.14 & 69.13 & 78.99 & 420.88 \\
          & NegFull$\sim70M$  & \textbf{56.26} & 80.46 & \textbf{88.46} & \textbf{40.29} & 67.01 & 76.90 & 409.38 & \textbf{59.68} & \textbf{83.36} & \textbf{90.06} & \textbf{42.66} & \textbf{69.05} & \textbf{79.06} & \textbf{423.87} \\

    \rowcolor[rgb]{ 1.0,  0.925,  0.905}  & None  & 54.26 & 79.20 & 87.20 & 39.21 & 66.19 & 76.49 & $402.55{{\uparrow_{10.8}}}$ & 56.26 & 80.26 & 88.04 & 40.59 & 67.81 & 78.08 & 411.04 \\
    \rowcolor[rgb]{ 1.0,  0.925,  0.905}     \multirow{-1.8}[0]{*}{+NEAT}  & CC12M$\sim12M$ & \textbf{56.26} & \textbf{80.94} & 88.38 & 40.21 & \textbf{67.36} & \textbf{77.53} & $\textbf{410.68}{{\uparrow_{10.2}}}$ & 58.60 & 82.32 & 89.42 & 42.15 & 69.01 & 78.86 & 420.36 \\
     \hline
    \multicolumn{16}{c}{\textit{Evaluation Dataset: MSR-VTT}} \\
    \hline
    \multirow{3}[1]{*}{CLIP} & None  & 20.30 & 39.80 & 49.90 & 23.90 & 45.80 & 56.70 & 236.40 & 25.50 & 48.40 & 60.90 & 28.00 & 50.60 & 60.50 & 273.90 \\
          & CC12M$\sim12M$ & 21.70 & 43.70 & 55.60 & 23.80 & 48.00 & 58.70 & 251.50 & 25.20 & 46.80 & 57.40 & 29.10 & 52.30 & 63.80 & 274.60 \\
          & NegFull$\sim70M$ & \textbf{25.20} & \textbf{50.10} & \textbf{59.50} & 20.60 & 43.90 & 54.60 & 253.90 & \textbf{29.00} & \textbf{52.00} & \textbf{63.90} & 23.70 & 46.90 & 57.30 & 272.80 \\

    \rowcolor[rgb]{ 1.0,  0.925,  0.905}  & None  & 24.10 & 44.10 & 54.20 & \textbf{24.80} & 47.60 & 58.10 & $252.90{{\uparrow_{16.5}}}$ & 26.10 & 48.10 & 59.50 & \textbf{28.80} & 50.00 & 58.80 & 271.30 \\
    \rowcolor[rgb]{ 1.0,  0.925,  0.905}    \multirow{-1.8}[0]{*}{+NEAT}     & CC12M$\sim12M$ & 22.70 & 45.30 & 56.90 & 24.50 & \textbf{48.90} & \textbf{60.20} & $\textbf{258.50}{{\uparrow_{7.0}}}$ & 25.40 & 47.90 & 58.90 & 28.60 & \textbf{51.10} & \textbf{62.80} & \textbf{274.70} \\
        \hdashline[3pt/3pt]
    ConCLIP & CC-Neg$\sim0.23M$ & 22.10 & 43.80 & 54.70 & 18.60 & 40.00 & 48.60 & 227.80 & 23.40 & 44.00 & 54.70 & 25.60 & 48.20 & 59.40 & 255.30 \\
             \hdashline[3pt/3pt]
    \multirow{3}[0]{*}{NegCLIP} & None  & 19.80 & 41.00 & 52.90 & 25.10 & 51.00 & 61.70 & 251.50 & 24.00 & 48.20 & 59.30 & 29.80 & 53.70 & 64.80 & 279.80 \\
          & CC12M$\sim12M$ & 23.30 & 46.50 & 58.50 & 27.00 & 53.60 & \textbf{64.10} & 273.00 & 27.80 & 51.80 & 63.80 & 30.00 & 55.50 & \textbf{65.50} & 294.40 \\
          & NegFull$\sim70M$  & \textbf{28.60} & \textbf{50.70} & \textbf{61.20} & \textbf{27.40} & 51.60 & 63.00 & \textbf{282.50} & \textbf{31.20} & \textbf{53.90} & \textbf{65.20} & \textbf{31.00} & \textbf{54.30} & 64.70 & \textbf{300.30} \\
          
    \rowcolor[rgb]{ 1.0,  0.925,  0.905}  & None  & 23.10 & 46.20 & 55.70 & 26.60 & 50.80 & 61.10 & $263.50{{\uparrow_{12.0}}}$ & 25.50 & 49.50 & 60.30 & 28.70 & 51.70 & 63.10 & 278.80 \\
    \rowcolor[rgb]{ 1.0,  0.925,  0.905}    \multirow{-1.8}[0]{*}{+NEAT}     & CC12M$\sim12M$ & 25.10 & 48.30 & 58.90 & 27.30 & \textbf{54.10} & 63.50 & $277.20{{\uparrow_{4.2}}}$  & 28.40 & 53.10 & 63.90 & 29.20 & 54.00 & 65.20 & 293.80 \\
\bottomrule
    \end{tabular}%
    }

      \label{tab: comparison_FT_methods}%
\end{table*}%
For clarity, we summarize datasets and tasks for negation understanding in \Cref{tab: datasets}, and we also show
some cases in Fig. \ref{fig: case_study}. In brief, the original validation set of MS-COCO \cite{lin2014microsoft} contains 5,000 images, where each is described by five captions. We adopt the LLM-rephrased variants from NegBench \cite{alhamoud2025vision}, comprising 5,000 Retrieval-Neg samples and 5,914 MCQ-Neg samples. VOC2007 \cite{everingham2010pascal} is an image dataset with 20 visual objects. We use the LLM-generated dataset from NegBench that contains 5,032 MCQ-Neg samples. The original test set of MSR-VTT \cite{xu2016msr} contains 2,990 video-caption pairs, where each video is captioned with 20 different descriptions. We adopt the LLM-rephrased variants from NegBench, comprising 1,000 Retrieval-Neg samples and 1,000 MCQ-Neg samples. Note that each video in Retrieval-Neg only has one corresponding caption. CheXpert \cite{irvin2019chexpert} is a large dataset of chest X-rays, where the validation set has 234 images with expert radiologist annotations. Since each image contains multiple disease labels, we select 4 representative diseases, \ie, Atelectasis, Cardiomegaly, Consolidation, and Lung Opacity, to create 616 negation classification pairs and 2,352 affirmative classification pairs as the base comparison. Moreover, we evaluate the zero-shot transfer ability for negation understanding on 9 widely used image datasets, including CIFAR10 \cite{krizhevsky2009learning}, CIFAR100 \cite{krizhevsky2009learning}, Caltch101 \cite{fei2006one}, OxfordPets \cite{parkhi2012cats}, RESISC45 \cite{cheng2017remote}, EuroSAT \cite{helber2019eurosat}, STL10 \cite{coates2011analysis}, SUN397 \cite{xiao2010sun}, and ImageNet1K \cite{deng2009imagenet}.

\subsubsection{Implementation Details} NEAT is a general TTA framework that could enhance most off-the-shelf pre-trained VLMs with negation understanding capabilities. Therefore, we select OpenAI CLIP \cite{radford2021learning}, CoNCLIP \cite{singh2024learn}, and NegCLIP \cite{yuksekgonul2022and} as the source models, where CoNCLIP and NegCLIP are two strong baselines designed to handle negation queries. To demonstrate the effectiveness of our method, we compare our NEAT with both post-training methods and state-of-the-art test-time adaptation methods. Following \cite{li2024test}, NEAT updates the parameters within the Layer Normalization (LN) layers in the textual encoder $f_{\theta_{T}}$ using the AdamW optimizer. All TTA baselines are conducted with a batch size of 256. The temperatures are set as $\tau_1$ = 0.03/0.07 and $\tau_2$ = 0.07/1.0 for image/video tasks, respectively. The balanced parameter $\lambda$ is set to 5 for all experiments. The trade-off parameter $\alpha$ is set to 1 for all experiments. To highlight practical adaptation during inference, we benchmark with offline and online updates to handle different scenarios. 



\begin{table*}[t]
  \centering
    \setlength{\aboverulesep}{-0.58pt}
\setlength{\belowrulesep}{0pt}
\setlength{\extrarowheight}{0pt}
  \renewcommand\arraystretch{1}
      \caption{Comparisons with state-of-the-art online TTA methods on MS-COCO and MSR-VTT datasets under negated queries. All methods are based on the ViT-B/32 CLIP. The best results are marked in \textbf{bold}.}
  \resizebox{0.92\textwidth}{!}{
    \begin{tabular}{l|cccccc|l|ccc|l}
    \toprule
    \multicolumn{1}{l|}{\multirow{1.3}[6]{*}{Method}} & \multicolumn{7}{c|}{\textbf{Retrieval-Neg Task}}               & \multicolumn{4}{c}{\textbf{Zero-shot MCQ-Neg Task}} \\
\cmidrule{2-12}    \multicolumn{1}{l|}{} & \multicolumn{3}{c|}{Image-to-Text} & \multicolumn{3}{c|}{Text-to-Image} & \multirow{1}[4]{*}{\ \ \ \ rSum} & \multicolumn{3}{c|}{MCQ Type} & \multirow{1}[4]{*}{\ Total Acc} \\
\cmidrule{2-7}\cmidrule{9-11}    \multicolumn{1}{l|}{} & R@1   & R@5   & \multicolumn{1}{c|}{R@10} & R@1   & R@5   & \multicolumn{1}{c|}{R@10} &       & Affirmed & Negated & \multicolumn{1}{c|}{Hybrid\ } &  \\
    \midrule
    \multicolumn{12}{c}{\textit{Evaluation Dataset: MS-COCO}} \\
    \midrule
    CLIP  & 45.30 & 69.78 & 79.10 & 24.96 & 47.93 & 59.40 & 326.47 & 69.09 & 6.84 & 39.21 & 39.25 \\
    *TENT & 44.74 & 70.44 & 79.94 & 25.24 & 49.24 & 60.88 & $330.48{{\uparrow_{4.01}}}$ & 45.23 & 15.72 & 17.40 & $26.43{{\downarrow_{12.82}}}$ \\
    *SAR  & 45.00 & 70.28 & 79.86 & 25.19 & 48.93 & 60.63 & $329.89{{\uparrow_{3.42}}}$  & 45.96 & 16.36 & 17.79 & $27.02{{\downarrow_{12.23}}}$ \\
    *READ & 43.44 & 68.04 & 77.54 & 24.93 & 47.61 & 59.23 & $320.79{{\downarrow_{5.68}}}$ & \textbf{68.85} & 7.54  & 38.87 & $39.26{{\uparrow_{0.01}}}$ \\
    *COME & 44.80 & 69.42 & 79.28 & 25.47 & 48.56 & 60.40 & $327.93{{\uparrow_{1.46}}}$ & 64.42 & 7.11  & 37.97 & $37.30{{\downarrow_{1.95}}}$ \\
    *TCR  & 44.88 & 70.14 & 79.40 & 27.50 & 51.89 & 63.18 & $336.99{{\uparrow_{10.52}}}$ & 45.37 & 15.56 & 21.17 & $27.71{{\downarrow_{11.54}}}$ \\
    \rowcolor[rgb]{ 1.0,  0.925,  0.905} *NEAT & \textbf{46.58} & \textbf{72.20} & \textbf{80.80} & \textbf{28.12} & \textbf{52.08} & \textbf{63.69} & $\textbf{343.47}{{\uparrow_{17.00}}}$ & 64.94 & \textbf{29.30} & \textbf{49.45} & $\textbf{48.41}{{\uparrow_{9.16}}}$ \\
    \midrule
    \multicolumn{12}{c}{\textit{Evaluation Dataset: MSR-VTT}} \\
    \midrule
    CLIP  & 20.30 & 39.80 & 49.90 & 23.90 & 45.80 & 56.70 & 236.40 & 62.39 & 13.31 & 20.83 & 32.10 \\
    *TENT & 19.90 & 42.50 & 52.60 & 25.00 & 46.20 & 57.30 & $243.50{{\uparrow_{7.1}}}$  & 53.43 & 11.33 & 17.95 & $27.50{{\downarrow_{4.6}}}$ \\
    *SAR  & 19.00 & 41.30 & 52.30 & 25.40 & 45.90 & 57.20 & $241.10{{\uparrow_{4.7}}}$  & 50.45 & 11.33 & 17.63 & $26.40{{\downarrow_{5.7}}}$ \\
    *READ & 20.40 & 40.20 & 51.80 & 23.40 & 45.70 & 56.30 & $237.80{{\uparrow_{1.4}}}$ & 63.28 & 13.31 & 19.87 & $32.10{{\uparrow_{0.0}}}$ \\
    *COME & 20.20 & 40.70 & 50.90 & 24.90 & 46.40 & 57.30 & $240.40{{\uparrow_{4.0}}}$ & 56.72 & 13.03 & 18.59 & $29.40{{\downarrow_{2.7}}}$ \\
    *TCR  & 19.70 & 41.90 & 51.10 & \textbf{25.70} & 46.80 & \textbf{58.30} & $243.50{{\uparrow_{7.1}}}$ & 55.82 & 13.60 & 19.55 & $29.60{{\downarrow_{2.5}}}$ \\
    \rowcolor[rgb]{ 1.0,  0.925,  0.905} *NEAT & \textbf{22.50} & \textbf{43.10} & \textbf{53.60} & 24.90 & \textbf{47.50} & 57.30 & $\textbf{248.90}{{\uparrow_{12.5}}}$ & \textbf{73.43} & \textbf{15.86} & \textbf{31.41} & $\textbf{40.00}{{\uparrow_{7.9}}}$ \\
    \bottomrule
    \end{tabular}%
}

      \label{tab: comparison_TTA_CLIP}%
\end{table*}%

\begin{table*}[t]
  \centering
    \setlength{\aboverulesep}{-0.58pt}
\setlength{\belowrulesep}{0pt}
\setlength{\extrarowheight}{0pt}
  \renewcommand\arraystretch{1}
      \caption{Comparisons with state-of-the-art online TTA methods on MS-COCO and MSR-VTT datasets under negated queries. All methods are based on the ViT-B/32 NegCLIP. The best results are marked in \textbf{bold}.}
  \resizebox{0.92\textwidth}{!}{
    \begin{tabular}{l|cccccc|l|ccc|l}
    \toprule
    \multicolumn{1}{l|}{\multirow{1.3}[6]{*}{Method}} & \multicolumn{7}{c|}{\textbf{Retrieval-Neg Task}}               & \multicolumn{4}{c}{\textbf{Zero-shot MCQ-Neg Task}} \\
\cmidrule{2-12}          & \multicolumn{3}{c|}{Image-to-Text} & \multicolumn{3}{c|}{Text-to-Image} & \multirow{1}[4]{*}{\ \ \ \ rSum} & \multicolumn{3}{c|}{MCQ Type} & \multirow{1}[4]{*}{\ Total Acc} \\
\cmidrule{2-7}\cmidrule{9-11}          & R@1   & R@5   & \multicolumn{1}{c|}{R@10} & R@1   & R@5   & R@10  &       & Affirmed & Negated & {Hybrid\ } &  \\
    \midrule
    \multicolumn{12}{c}{\textit{Evaluation Dataset: MS-COCO}} \\
    \midrule
    NegCLIP & 51.64 & 77.92 & 86.06 & 37.02 & 64.39 & 74.77 & 391.80 & 51.67 & 16.15 & 17.15 & 28.69 \\
    *TENT & 53.74 & 79.70 & 88.24 & 38.65 & 66.55 & 77.20 & $404.08{{\uparrow_{12.28}}}$ & 41.78 & 17.86 & 8.80  & $23.00{{\downarrow_{5.69}}}$ \\
    *SAR  & 53.36 & 79.38 & 87.86 & 39.18 & 66.95 & 77.27 & $404.00{{\uparrow_{12.20}}}$  & 41.34 & 19.30 & 8.75  & $23.28{{\downarrow_{5.41}}}$ \\
    *READ & 52.30 & 78.66 & 87.00 & 37.32 & 65.10 & 76.06 & $396.44{{\uparrow_{4.64}}}$  & 51.82 & 16.15 & 17.25 & $28.78{{\uparrow_{0.09}}}$ \\
    *COME & 51.38 & 77.42 & 86.08 & 37.25 & 65.10 & 76.11 & $393.34{{\uparrow_{1.54}}}$  & 44.34 & 15.40 & 12.28 & $24.28{{\downarrow_{4.41}}}$ \\
    *TCR  & \textbf{54.02} & \textbf{79.98} & \textbf{88.32} & 39.01 & \textbf{67.00} & 77.54 & $\textbf{405.87}{{\uparrow_{14.07}}}$  & 41.04 & 18.50 & 8.75  & $22.93{{\downarrow_{5.76}}}$ \\
    \rowcolor[rgb]{ 1.0,  0.925,  0.905} *NEAT & 53.68 & 79.40 & 88.02 & \textbf{39.09} & 66.99 & \textbf{77.78} & $404.96{{\uparrow_{13.16}}}$  & \textbf{52.12} & \textbf{28.45} & \textbf{34.79} & $\textbf{38.74}{{\uparrow_{10.05}}}$ \\
    \midrule
    \multicolumn{12}{c}{\textit{Evaluation Dataset: MSR-VTT}} \\
    \midrule
    NegCLIP & 19.80 & 41.00 & 52.90 & 25.10 & 51.00 & 61.70 & 251.50 & 51.94 & 12.46 & 17.63 & 27.30 \\
    *TENT & 19.90 & 41.50 & 52.00 & 26.10 & 51.10 & 61.50 & $252.10{{\uparrow_{0.6}}}$  & 41.19 & 10.20 & 11.54 & $21.00{{\downarrow_{6.3}}}$  \\
    *SAR  & 20.40 & 43.00 & 53.70 & 25.90 & 51.30 & 61.90 & $256.20{{\uparrow_{4.7}}}$  & 43.58 & 11.33 & 12.50 & $22.50{{\downarrow_{4.8}}}$  \\
    *READ & 21.40 & 42.90 & 54.00 & 25.00 & 51.20 & 61.40 & $255.90{{\uparrow_{4.4}}}$  & 54.03 & 12.46 & 18.91 & $28.40{{\uparrow_{1.1}}}$  \\
    *COME & 21.30 & 42.50 & 54.40 & 25.10 & 51.30 & 62.10 & $256.70{{\uparrow_{5.2}}}$  & 49.85 & 11.61 & 17.31 & $26.20{{\downarrow_{1.1}}}$  \\
    *TCR  & 21.20 & 42.80 & 54.20 & 25.40 & 51.40 & 62.30 & $257.30{{\uparrow_{5.8}}}$  & 48.66 & 12.75 & 17.31 & $26.20{{\downarrow_{1.1}}}$  \\
    \rowcolor[rgb]{ 1.0,  0.925,  0.905} *NEAT & \textbf{23.80} & \textbf{45.30} & \textbf{56.80} & \textbf{27.20} & \textbf{52.50} & \textbf{62.80} & $\textbf{268.40}{{\uparrow_{16.9}}}$  & \textbf{69.85} & \textbf{15.86} & \textbf{27.88} & $\textbf{37.70}{{\uparrow_{10.4}}}$  \\
    \bottomrule
    \end{tabular}%
}

      \label{tab: comparison_TTA_NegCLIP}%
\end{table*}%

\begin{table*}[t]
  \centering
    \setlength{\aboverulesep}{-0.58pt}
\setlength{\belowrulesep}{0pt}
\setlength{\extrarowheight}{0pt}
  \renewcommand\arraystretch{1}
      \caption{Comparisons with state-of-the-art offline TTA methods on MS-COCO datasets under negated queries. All methods are based on the ViT-B/16 BLIP. The best results are marked in \textbf{bold}.}
  \resizebox{0.92\textwidth}{!}{
    \begin{tabular}{l|cccccc|l|ccc|l}
    \toprule
    \multirow{1.3}[6]{*}{Method} & \multicolumn{7}{c|}{\textbf{Retrieval Neg Task}}      & \multicolumn{4}{c}{\textbf{Zero-shot MCQ-Neg Task}} \\
\cmidrule{2-12}          & \multicolumn{3}{c|}{Image-to-Text} & \multicolumn{3}{c|}{Text-to-Image} & \multirow{1}[4]{*}{rSum} & \multicolumn{3}{c|}{MCQ Type} & \multirow{1}[4]{*}{Total} \\
\cmidrule{2-7}\cmidrule{9-11}          & R@1   & R@5   & \multicolumn{1}{c|}{R@10} & R@1   & R@5   & R@10  &       & Affirmed & Negated & Hybrid &  \\
    \midrule
    \multicolumn{12}{c}{\textit{Base model: ViT-B/16 BLIP}} \\
    \midrule
    BLIP  & 52.60 & 77.90 & 85.76 & 33.58 & 58.96 & 69.68 & 378.48 & 33.81 & 16.68 & 5.12  & 18.63 \\
    *TENT & 37.96 & 66.98 & 79.50 & 40.54 & 67.16 & 78.33 & $370.47{{\downarrow_{8.01}}}$ & 19.14 & 20.70 & 2.68  & $14.03{{\downarrow_{4.60}}}$ \\
    *SAR  & 37.24 & 67.74 & 78.94 & 40.35 & 68.06 & 78.04 & $370.37{{\downarrow_{8.11}}}$ & 11.71 & 23.53 & 1.09  & $11.84{{\downarrow_{6.79}}}$ \\
    *READ & 55.24 & 79.56 & 87.36 & 36.32 & 62.90 & 73.27 & $394.65{{\uparrow_{16.17}}}$ & 35.73 & 16.52 & 5.62  & $19.41{{\uparrow_{0.78}}}$ \\
    *COME & 44.84 & 70.80 & 79.64 & 37.11 & 63.78 & 74.17 & $370.34{{\downarrow_{8.14}}}$ & 31.74 & 17.59 & 3.68  & $17.72{{\downarrow_{0.91}}}$ \\
    *TCR  & 48.38 & 76.10 & 85.34 & \textbf{46.10} & \textbf{73.26} & \textbf{82.39} & $411.57{{\uparrow_{33.09}}}$ & 20.62 & 31.76 & 5.47  & $18.99{{\uparrow_{0.36}}}$ \\
    \rowcolor[rgb]{ 1.0,  0.925,  0.905} *NEAT & \textbf{59.22} & \textbf{83.00} & \textbf{89.98} & 41.12 & 68.40 & 78.47 & $\textbf{420.19}{{\uparrow_{41.71}}}$  & \textbf{77.21} & \textbf{28.34} & \textbf{49.95} & $\textbf{52.49}{{\uparrow_{33.86}}}$ \\
    \midrule
    \multicolumn{12}{c}{\textit{Base model: ViT-B/16 BLIP Fine-tuned on MS-COCO Training Data}} \\
    \midrule
    BLIP  & 69.98 & 90.98 & 95.00 & 54.69 & 79.86 & 87.34 & 477.85 & 50.69 & 12.94 & 16.65 & 27.17 \\
    *TENT & 68.14 & 89.46 & 94.10 & 57.44 & 82.02 & 88.85 & $480.01{{\uparrow_{2.16}}}$ & 47.79 & 19.36 & 28.28 & $32.16{{\uparrow_{4.99}}}$ \\
    *SAR  & 66.40 & 88.10 & 93.50 & 57.51 & 82.13 & 88.90 & $476.54{{\downarrow_{1.31}}}$ & 47.29 & 19.36 & 29.97 & $32.57{{\uparrow_{5.40}}}$ \\
    *READ & 70.40 & 91.28 & 95.22 & 53.01 & 78.62 & 86.34 & $474.87{{\downarrow_{2.98}}}$ & 53.00 & 11.76 & 17.20 & $27.78{{\uparrow_{0.61}}}$ \\
    *COME & 69.38 & 90.62 & 94.96 & 57.63 & 81.81 & 88.75 & $483.15{{\uparrow_{5.30}}}$ & 55.81 & 11.12 & 16.90 & $28.44{{\uparrow_{1.27}}}$ \\
    *TCR  & 68.58 & 89.66 & 94.12 & \textbf{58.78} & \textbf{83.10} & \textbf{89.66} & $483.90{{\uparrow_{6.05}}}$ & 47.98 & \textbf{19.52} & 30.07 & $32.89{{\uparrow_{5.72}}}$ \\
    \rowcolor[rgb]{ 1.0,  0.925,  0.905} *NEAT & \textbf{74.18} & \textbf{92.40} & \textbf{95.98} & 56.40 & 81.11 & 88.47 & $\textbf{488.54}{{\uparrow_{10.69}}}$ & \textbf{58.27} & 18.29 & \textbf{46.22} & $\textbf{41.53}{{\uparrow_{14.36}}}$ \\
    \bottomrule
    \end{tabular}%
}
      \label{tab: comparison_TTA_BLIP}%
\end{table*}%

\subsection{Comparisons with State of the Arts}
\subsubsection{Compared to Post-training Methods}
In this section, we compare our NEAT with post-training methods that fine-tune VLMs on labeled negation-enhanced datasets. Specifically, CC-Neg \cite{singh2024learn} is a synthetic dataset of 0.23 million image-caption pairs along with high-quality negated captions as distractors. NegFull \cite{alhamoud2025vision} is a large-scale dataset of 70 million pairs generated from CC12M \cite{changpinyo2021conceptual}, where each image is captioned with multiple incorporated negated objects and hard negatives based annotations. For fair comparison with offline baselines, our NEAT applies offline adaptation where the model is first updated on all available data before proceeding to inference. In \Cref{tab: comparison_FT_methods}, we present the cross-modal retrieval performance on MS-COCO and MSR-VTT datasets with negation queries. Note that our offline adaptation is conducted only on unlabeled retrieval-Neg data, and we also report the performance on normal retrieval tasks after TTA or post-training. From the results, we could observe the following conclusions:

\begin{itemize}
\item Despite being trained on massive data, CLIP suffers a 33.07\% drop in terms of the sum in retrieval (\ie, rSum) on MS-COCO and a 37.50\% drop on MSR-VTT, showing that VLMs have difficulty in handling negation queries. 

\item Our NEAT could significantly improve the negation comprehension ability of pre-trained VLMs, \eg, it improves the rSum of CLIP by 11.80\%$\sim$29.54\% on MS-COCO. 

\item Our NEAT could generalize to different VLMs and modalities. For instance, it improves the rSum of NegCLIP and its CC12M fine-tuned model by 10.8\% and 10.2\% on MS-COCO, respectively. In addition, although only adapted on 1,000 video-text pairs, NEAT still improves the rSum of CLIP by 7.0\%$\sim$16.5\% on MSR-VTT. 

\item The normalization layers modified by NEAT do not impair VLMs' ability for normal vision-language comprehension on affirmative statements, with rSum performance remaining within stable and acceptable ranges, \ie. $-4.75\%\sim+4.91\%$. 

\item Remarkably, our NEAT achieves comparable or even superior performance to post-trained methods while avoiding the demand for massive negation data. Compared to the SOTA method that fine-tunes on 70 million well-designed negation-enriched pairs, \ie. NegFull, our NEAT surpasses its CLIP variant by a clear margin on MS-COCO, achieving much to 25.51\% absolute improvement in rSum performance. 
\end{itemize}

\begin{table*}[t]
  \centering
    \setlength{\aboverulesep}{-0.58pt}
\setlength{\belowrulesep}{0pt}
\setlength{\extrarowheight}{0pt}
  \renewcommand\arraystretch{1}
      \caption{Zero-shot transfer evaluation on two template-based negation classification tasks. 
    The best results are marked in \textbf{bold}.}
  \resizebox{1\textwidth}{!}{
    \begin{tabular}{l|ccccccccc|c}
    \toprule
    Method & CIFAR10 & CIFAR100 & Caltch101 & OxfordPets & RESISC45 & EuroSAT & STL10 & SUN397 & ImageNeg1K & AVERAGE \\
    \midrule
    \multicolumn{11}{c}{\textit{Negation-conditioned Zero-shot Classification: Top-1 Accuracy}$\uparrow$} \\
    \midrule
    CLIP  & 64.44 & 49.30 & 75.00 & 61.59 & 44.75 & 25.13 & 77.41 & 47.56 & 45.97 & 54.57 \\
    NegCLIP & 66.13 & 44.51 & 61.89 & 59.80 & 30.59 & 26.39 & 79.60 & 33.99 & 35.30 & 48.69 \\
    ConCLIP & 74.95 & 19.84 & 22.00 & 26.79 & 15.05 & 15.04 & 89.79 & 26.42 & 18.06 & 34.22 \\
    NegCLIP+NegFull FT & \textbf{88.40} & 56.68 & \textbf{79.29} & 67.70 & \textbf{52.73} & \textbf{38.94} & 95.16 & \textbf{53.09} & \textbf{49.74} & \textbf{64.64} \\
    CLIP+TCR & 66.74 & 49.22 & 75.49 & 60.75 & 36.92 & 37.87 & 78.03 & 43.35 & 39.51 & 54.21 \\
    \rowcolor[rgb]{ 1.0,  0.925,  0.905} CLIP+NEAT & 87.22 & \textbf{57.96} & 74.34 & \textbf{71.90} & 49.62 & 33.24 & \textbf{95.59} & 50.87 & 47.90 & 63.18 \\
    \midrule
    \multicolumn{11}{c}{\textit{Reversed Negation-conditioned Zero-shot Classification: Top-1 Error Rate}$\downarrow$} \\
    \midrule
    CLIP  & 22.83 & 8.29  & 16.07 & 5.70  & 12.13 & 9.80  & 19.13 & 3.81  & 5.01  & 11.42 \\
    NegCLIP & 23.25 & 10.95 & 23.38 & 10.66 & 15.38 & 12.37 & 17.70 & 9.29  & 7.55  & 14.50 \\
    ConCLIP & 9.97  & 1.72  & 2.86  & 4.42  & 1.62  & 1.54  & 0.53  & 0.40  & 0.28  & 2.59 \\
    NegCLIP+NegFull FT & 2.47  & 1.31  & 1.53  & 0.74  & 4.22  & 8.94  & 0.34  & 0.19  & 0.17  & 2.21 \\
    CLIP+TCR & 14.44 & 7.81  & 12.46 & 10.58 & 13.27 & 19.19 & 12.76 & 6.65  & 7.54  & 11.63 \\
    \rowcolor[rgb]{ 1.0,  0.925,  0.905} CLIP+NEAT & \textbf{2.24} & \textbf{0.44} & \textbf{1.60} & \textbf{0.22} & \textbf{0.43} & \textbf{6.85} & \textbf{0.33} & \textbf{0.07} & \textbf{0.04} & \textbf{1.36} \\
    \bottomrule
    \end{tabular}%
}

      \label{tab: image_classification_results}%
\end{table*}%

\subsubsection{Compared to Test-time Adaptation Methods}
In this section, we compare our NEAT with five SOTA TTA methods, \ie, TENT \cite{wang2020tent}, SAR \cite{niutowards}, READ \cite{yang2024test}, COME, and TCR \cite{li2024test}, under negated queries. Among the baseline methods, Tent, SAR, and COME are unimodal TTA approaches based on the entropy-minimizing objective or its variants, while READ and TCR are multi-modal TTA approaches that further mitigate the modality bias. As entropy-based TTA methods are sensitive to the temperature parameter, we maintain the same $\tau_1$ value as NEAT for a fair comparison. All methods are evaluated through online adaptation, where LN layers are updated on the current batch and then make a prediction. In detail, all methods are adapted on two unlabeled Retrieval-Neg datasets, \ie, MS-COCO and MSR-VTT, and report the retrieval performance. To further investigate the generalization ability, we evaluate the classification accuracy of the adapted VLM on the corresponding MCQ-Neg task. As shown in \Cref{tab: comparison_TTA_CLIP} and \Cref{tab: comparison_TTA_NegCLIP}, one could conclude that NEAT remarkably boosts the negation understanding of the baselines. More specifically,

\begin{itemize}
\item Most existing TTA methods only achieve marginal improvements over the base model,  indicating that negated concepts pose challenging distribution shift problems.

\item Although some multi-modal methods show considerable performance gains on Retrieval-Neg Task, \ie, TCR, the accuracy on the corresponding MCQ-Neg task drops unexpectedly by $1.10\%$ to $11.54\%$. This indicates that the observed improvements may be attributed to reducing modality gaps rather than addressing the concept shift problem arising from negation understanding.

\item The extensions with NEAT achieve remarkably superior overall performance compared to all baselines under all settings. For example, on the MS-COCO dataset with the CLIP-based model, our NEAT surpasses the SOTA baseline by 6.48\% on the Retrieval-Neg task and 20.7\% on the more challenging zero-shot MCQ-Neg task. It also outperforms the SOTA baseline by 11.1\% on Retrieval-Neg and 11.5\% on MCQ-Neg for the MSR-VTT dataset with the NegCLIP-based model, indicating that our NEAT can generalize effectively across different scenarios.
\end{itemize}

In addition to the evaluation for CLIP-based VLMs, we also conduct experiments for BLIP-based VLMs \cite{li2022blip}, \ie, a base model pretrained on 129 million data and its fine-tuned model on the MS-COCO dataset. From Fig. \ref{tab: comparison_TTA_BLIP}, one could see that some TTA baselines show large performance variations between image-to-text and text-to-image retrieval after adaptation. This may be because BLIP uses multiple pretraining objectives while adaptation only focuses on the image-text contrastive one. The phenomenon is significantly alleviated when the model is fine-tuned on corresponding image-text retrieval tasks. Moreover, like the observations on CLIP-based models, our NEAT remarkably
boosts the effectiveness of the baselines. Specifically, NEAT improves BLIP by 41.71\% (rSum) for cross-modal retrieval and 33.86\% (accuracy) for multiple choice questions.

\subsection{Generalizability Analysis}
In this section, we investigate whether NEAT can learn generalizable negation patterns rather than simply fitting to the negation data. To this end, we conduct a series of studies of increasing difficulty to verify the performance of NEAT-adapted VLMs on unseen negation data and tasks. Unless otherwise stated, all TTA methods are adapted on the MS-COCO Retrieval-Neg dataset using the CLIP ViT-B/32 model.

\subsubsection{Easy: Cross-Dataset Generalization}
We first evaluate VLMs that were adapted by different TTA methods on the unseen VOC2007 MCQ-Neg task. As demonstrated in \Cref{tab: voc2007_results}, although some TTA baselines can adapt to negation data and enhance performance on such data, the adapted models fail to generalize to novel negation data, leading to substantial performance drops. In contrast, our NEAT remarkably improves CLIP from 38.70\% to 55.54\% and NegCLIP from 30.47\% to 49.73\% on the unseen voc2007 dataset.

\subsubsection{Normal: Cross-Task Generalization}
We further test NEAT-adapted VLMs across 9 widely used image classification datasets using two template-based negation understanding tasks. As elaborated in \ref{subsec:concept-shifts}, we randomly introduce distractor labels to construct the negation-conditioned caption, and swap its two class names to construct the reversed negation-conditioned caption. The comparison results are shown in \Cref{tab: image_classification_results}. From the table, one could see that most negation-aware methods cannot generalize to downstream image classification tasks, performing even worse than the base model. Additionally, post-training on millions of well-designed negation data can endow VLMs with strong negation generalization capabilities, but the performance remains sub-optimal for the more difficult reversed negation-conditioned task. Remarkably, although only updating LN layers with minimal unlabeled data, our approach not only improves the accuracy of CLIP from 54.57\% to 63.18\% (average) for the negation-conditioned task, but also reduces the error rate of CLIP from 11.42\% to 1.36\% (average) for the reversed negation-conditioned task.

\begin{figure*}[t]
\centering
\subfloat[\small Negation-separated Prediction]{\includegraphics[width=0.32\textwidth]{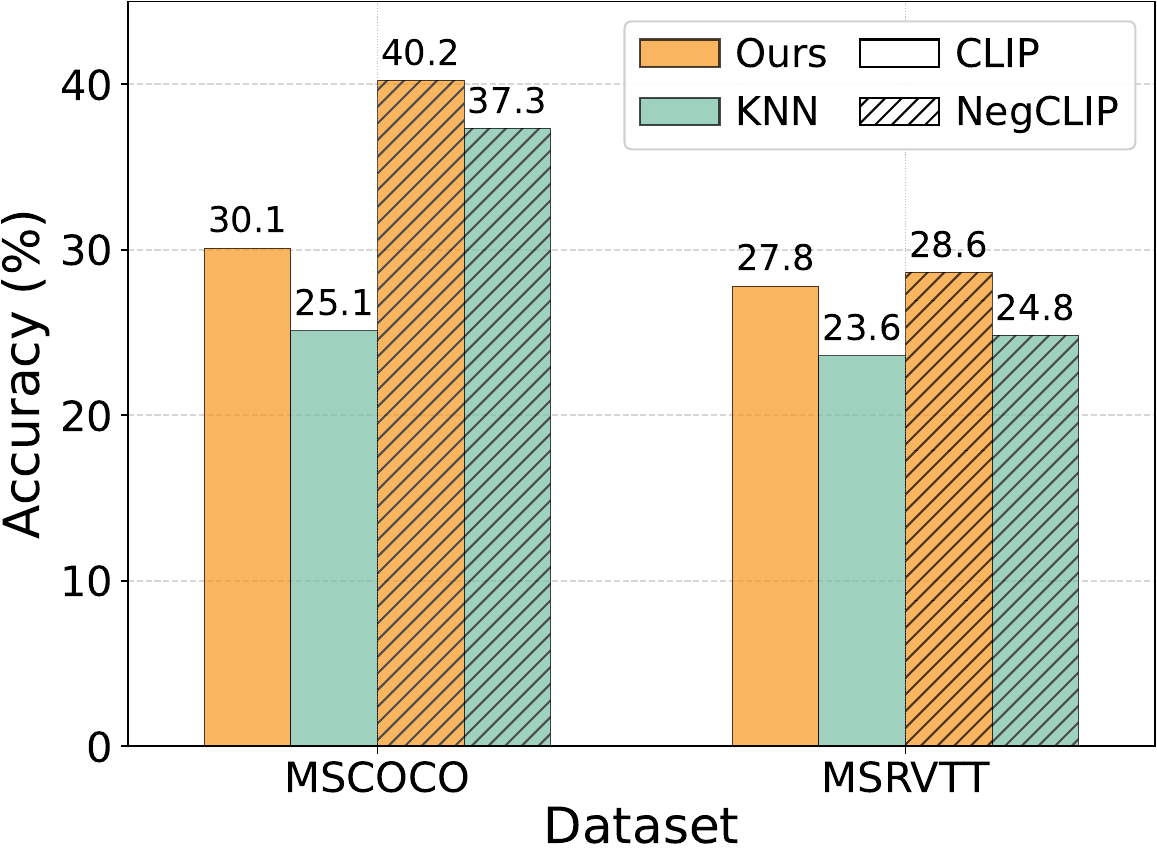}\label{fig: ablation_bar}}\hspace{0.15cm}
\subfloat[\small Parameter Ablation]{\includegraphics[width=0.32\textwidth]{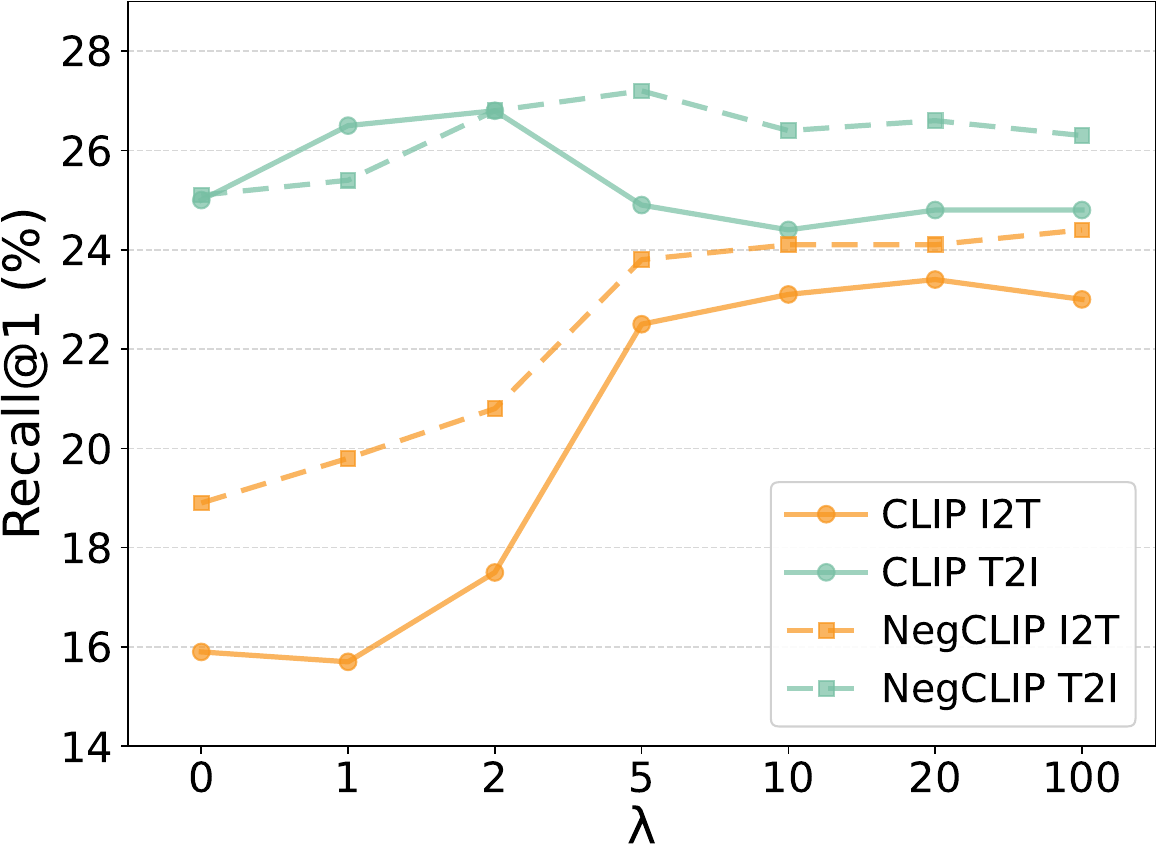}\label{fig: ablation_lambda}}\hspace{0.15cm}
\subfloat[\small Temperature Sensitivity]{\includegraphics[width=0.32\textwidth]{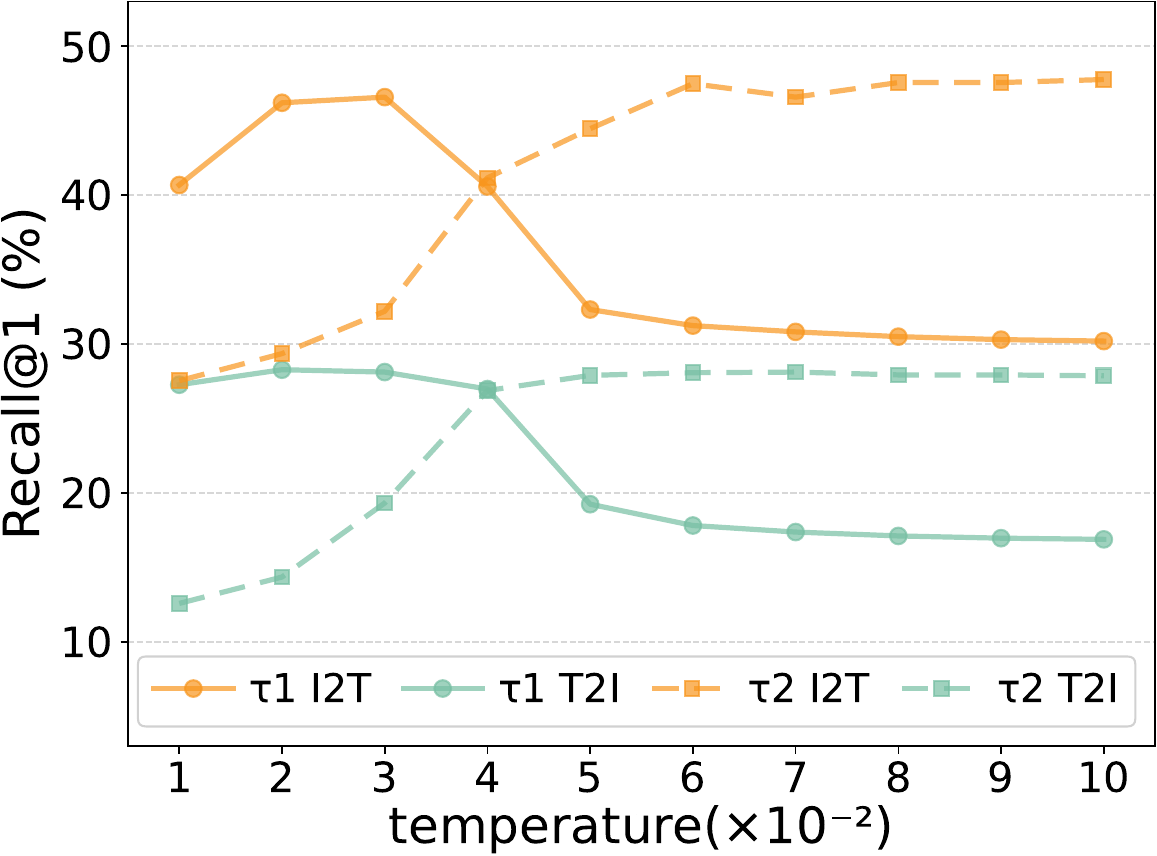}\label{fig: ablation_tau}}
\caption{Finer-grained ablation studies. (a) The top-1 prediction accuracy on the vanilla KNN-based prediction and our negation-separated prediction. (b) The parameter analysis of $\lambda$ on MSR-VTT. (c) The parameter analysis of $\tau_1$ and $\tau_2$ on MS-COCO.}
\label{fig: finer-grained ablation}
\end{figure*}

\begin{table}[t]
  \centering
    \setlength{\aboverulesep}{-0.58pt}
\setlength{\belowrulesep}{0pt}
\setlength{\extrarowheight}{0pt}
  \renewcommand\arraystretch{1}
      \caption{Zero-shot transfer evaluation of TTA methods on the VOC2007 MCQ-Neg task after adaptation on unlabeled MS-COCO Retrieval-Neg data.}
  \resizebox{0.95\columnwidth}{!}{
    \begin{tabular}{l|ccc|l}
    \toprule
    \multirow{1}[4]{*}{Method} & \multicolumn{3}{c|}{MCQ-Neg Type} & \multirow{1}[4]{*}{\ \ Total Acc} \\
\cmidrule{2-4}          & Affirmed & Negated & Hybrid &  \\
    \midrule
    CLIP  & 82.70 & 3.30  & 58.93 & 38.70 \\
    *TENT & 60.12 & 5.14  & 38.51 & $27.37{{\downarrow_{11.33}}}$ \\
    *SAR  & 61.14 & 5.09  & 39.59 & $27.97{{\downarrow_{10.73}}}$ \\
    *READ & 80.79 & 3.87  & 58.08 & $38.30{{\downarrow_{0.40}}}$ \\
    *COME & 79.33 & 2.92  & 59.07 & $38.14{{\downarrow_{0.56}}}$ \\
    *TCR  & 61.00 & 5.37  & 45.74 & $30.79{{\downarrow_{7.91}}}$ \\
    \rowcolor[rgb]{ 1.0,  0.925,  0.905} *NEAT & \textbf{81.09} & \textbf{23.86} & \textbf{77.87} & $\textbf{55.54}{{\uparrow_{16.84}}}$ \\
    \midrule
    NegCLIP & 75.37 & 5.79  & 40.31 & 30.47 \\
    *TENT & 68.48 & 6.03  & 31.10 & $25.60{{\downarrow_{4.87}}}$ \\
    *SAR  & 68.62 & 6.08  & 31.96 & $26.02{{\downarrow_{4.45}}}$ \\
    *READ & 76.25 & 5.70  & 41.07 & $30.93{{\uparrow_{0.46}}}$ \\
    *COME & 73.02 & 5.52  & 36.36 & $28.32{{\downarrow_{2.15}}}$ \\
    *TCR  & 68.62 & 6.13  & 32.05 & $26.08{{\downarrow_{4.39}}}$ \\
    \rowcolor[rgb]{ 1.0,  0.925,  0.905} *NEAT & \textbf{77.13} & \textbf{18.34} & \textbf{71.23} & $\textbf{49.73}{{\uparrow_{19.26}}}$ \\
    \bottomrule
    \end{tabular}%
}

      \label{tab: voc2007_results}%
\end{table}%

\begin{table}[t]
  \centering
    \setlength{\aboverulesep}{-0.58pt}
\setlength{\belowrulesep}{0pt}
\setlength{\extrarowheight}{0pt}
    \addtolength{\tabcolsep}{-2.5pt}  
  \renewcommand\arraystretch{1}
      \caption{Zero-shot transfer evaluation on CheXpert by replacing LN layers from different negation-aware models.}
  \resizebox{1\columnwidth}{!}{
    \begin{tabular}{l|cc|cc|cc|cc}
    \toprule
    \multirow{0.9}[4]{*}{Method} & \multicolumn{2}{c|}{Atelectasis} & \multicolumn{2}{c|}{Cardiomegaly} & \multicolumn{2}{c|}{Consolidation} & \multicolumn{2}{c}{Lung Opacity} \\
\cmidrule{2-9}          & Aff   & Neg   & Aff   & Neg   & Aff   & Neg   & Aff   & Neg \\
    \midrule
    PLIP  & 67.49 & \textbf{84.38} & 21.43 & 27.94 & 41.90 & \textbf{90.91} & 84.38 & 34.13 \\
    CONCLIP & 66.67 & 83.75 & 20.85 & 27.21 & 41.43 & \textbf{90.91} & 89.00 & 41.67 \\
    CLIP NegFull  & 66.34 & 83.75 & 20.85 & 27.21 & 41.43 & \textbf{90.91} & 84.97 & 34.13 \\
    CLIP TCR   & 75.08 & 58.75 & \textbf{44.40} & \textbf{44.12} & 36.67 & 77.27 & 84.58 & 42.46 \\
    \rowcolor[rgb]{ 1.0,  0.925,  0.905} CLIP NEAT  & \textbf{80.53} & \textbf{84.38} & 33.20 & 31.62 & \textbf{61.43} & 89.39 & \textbf{98.92} & \textbf{70.24} \\
    \bottomrule
    \end{tabular}%
}

      \label{tab: chexpert_results}%
\end{table}%

\subsubsection{Hard: Cross-Normalization Layers Generalization} Beyond validating on complete models, we further investigate whether the adapted LN layers can be directly applied to domain-specific models. To this end, we use the medical foundation VLM, \ie, PLIP \cite{huang2023visual}, as the base model, replacing its textual LN layers with those from negation-aware VLMs to observe the performance impact. Following \cite{alhamoud2025vision}, we conduct evaluations on two classification settings. Specifically, the Binary Classification-Aff task requires the model to correctly associate each image with the two affirmative statements, such as `This image shows Atelectasis' and `This image shows Consolidation'. The Binary Classification-Neg task requires the model to distinguish each image with the presence or absence of a medical condition, such as `This image shows Atelectasis' and `This image does not show Atelectasis'. As shown in \Cref{tab: chexpert_results}, one can observe that PLIP exhibits robust negation understanding abilities in most disease recognition tasks, \eg, Atelectasis and Consolidation, suggesting that its training medical reports likely contain explicit negation expressions. Second, using the LN layers from post-training methods does not lead to significant performance changes, indicating that their negation understanding capabilities are more attributable to the encoder or decoder layers. Third, using the LN layers adapted by TCR brings performance gains in certain scenarios, but exhibits instability in others. Notably, our NEAT provides stable performance improvements overall, especially for the most frequent \cite{alhamoud2025vision} condition Lung Opacity, \ie, achieving 14.54\% and 36.11\% improvements for Binary Classification-Aff and Classification-Neg tasks, respectively.

\subsection{Ablation Study} 
\subsubsection{Impact of Each Component}
To study the influence of specific components in our method, we first carry out ablation studies on the MS-COCO Retrieval-Neg data with different training objects. \Cref{tab: ablation} reports the performance of the Retrieval-neg and the MCQ-Neg tasks. From the results, we observe the following conclusions: 1) $\mathcal{L}_{ent}$ boosts performance through the negation-separated prediction, \ie, improving the R@1 of the base CLIP by 2.76\% and 4.24\% on text and image retrieval, respectively. However, naive entropy minimization fails to generalize the model to more difficult negation understanding scenarios, \eg, performance drops significantly on the MCQ task. 2) $\mathcal{L}_{sr}$ is crucial for learning generalizable negation understanding capabilities, increasing the MCQ accuracy of the $\mathcal{L}_{ent}$ variant from 28.69\% to 45.91\%. 3) NEAT achieves overall optimal performance when all the loss terms are employed, showing that all three components make contributions. In addition, we conduct experiments to verify the effectiveness of the proposed negation-separated prediction in \Cref{subsec: Refined P}. We compared with the KNN-based prediction \cite{li2024test} and report the zero-shot top-1 prediction accuracy. As shown in Fig. \ref{fig: ablation_bar}, our refinement strategy substantially outperforms the KNN variant, which signifies that the simple feature-driven prediction is insufficient for handling negation.

\begin{figure*}[t]
\centering
\subfloat[\small CLIP Normal]{\includegraphics[width=0.19\textwidth]{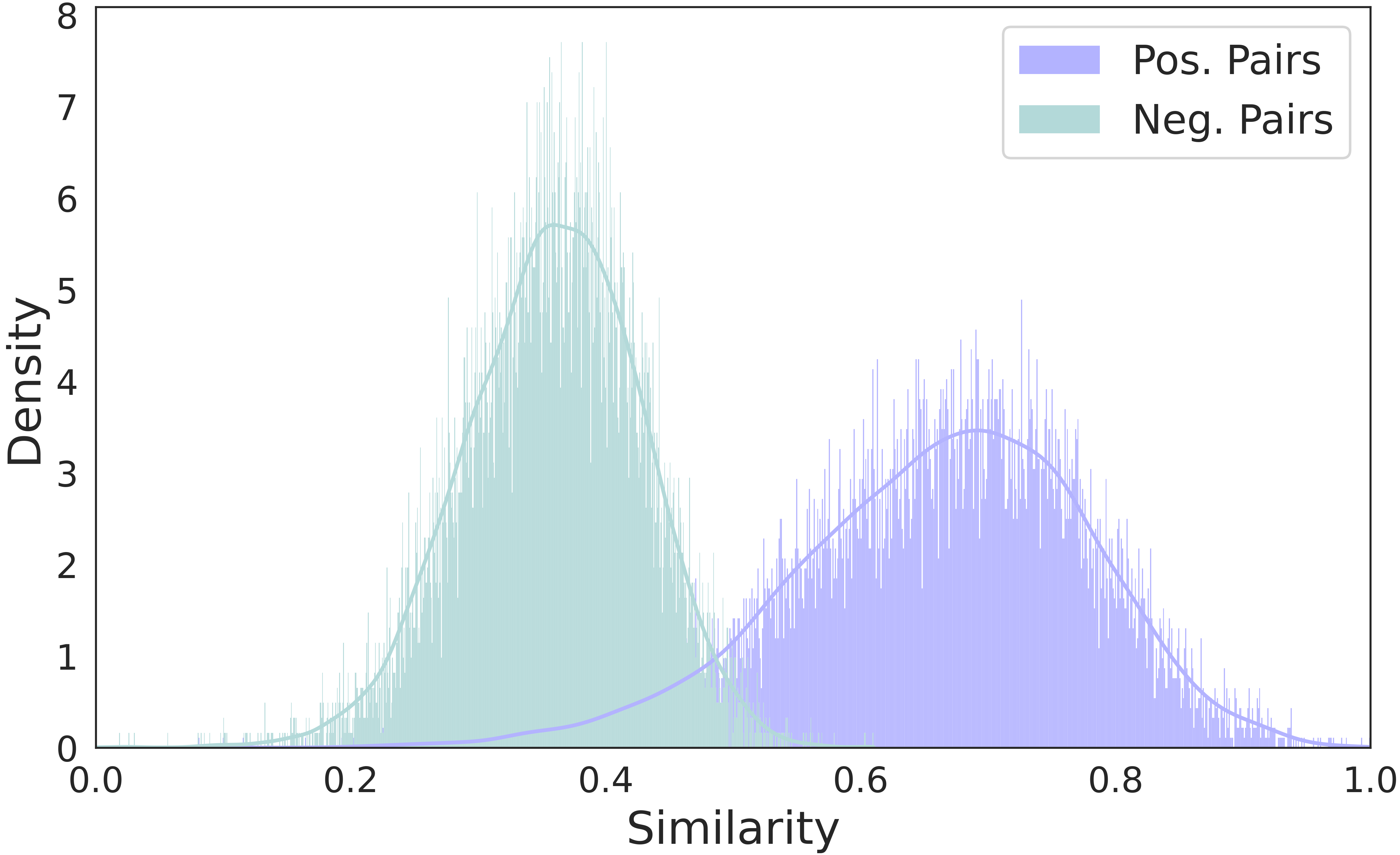}\label{fig: clip_sims_norm}}\hspace{0cm}
\subfloat[\small ConCLIP Normal]{\includegraphics[width=0.19\textwidth]{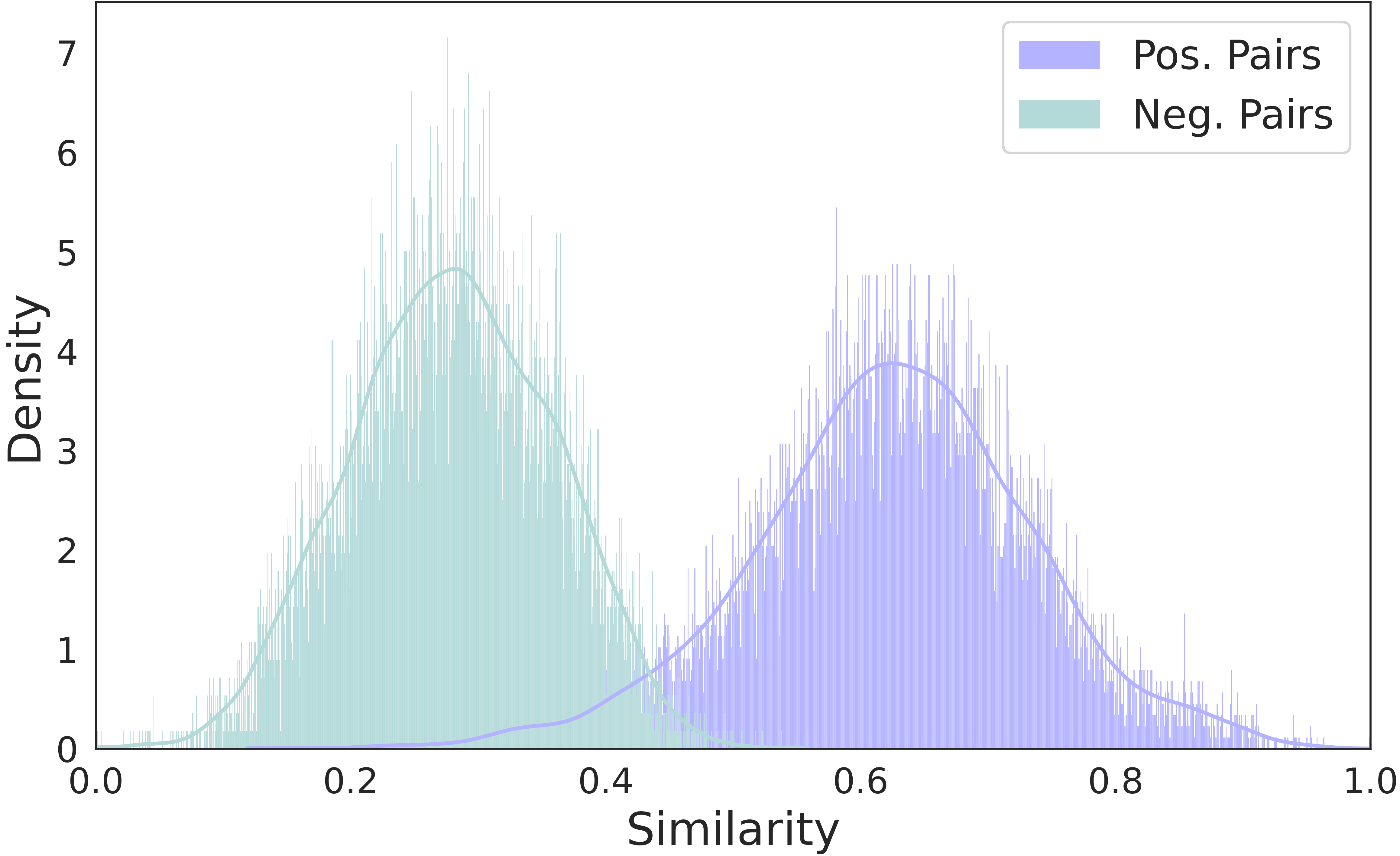}\label{fig: conclip_sims_norm}}\hspace{0cm}
\subfloat[\small NegFull Normal]{\includegraphics[width=0.19\textwidth]{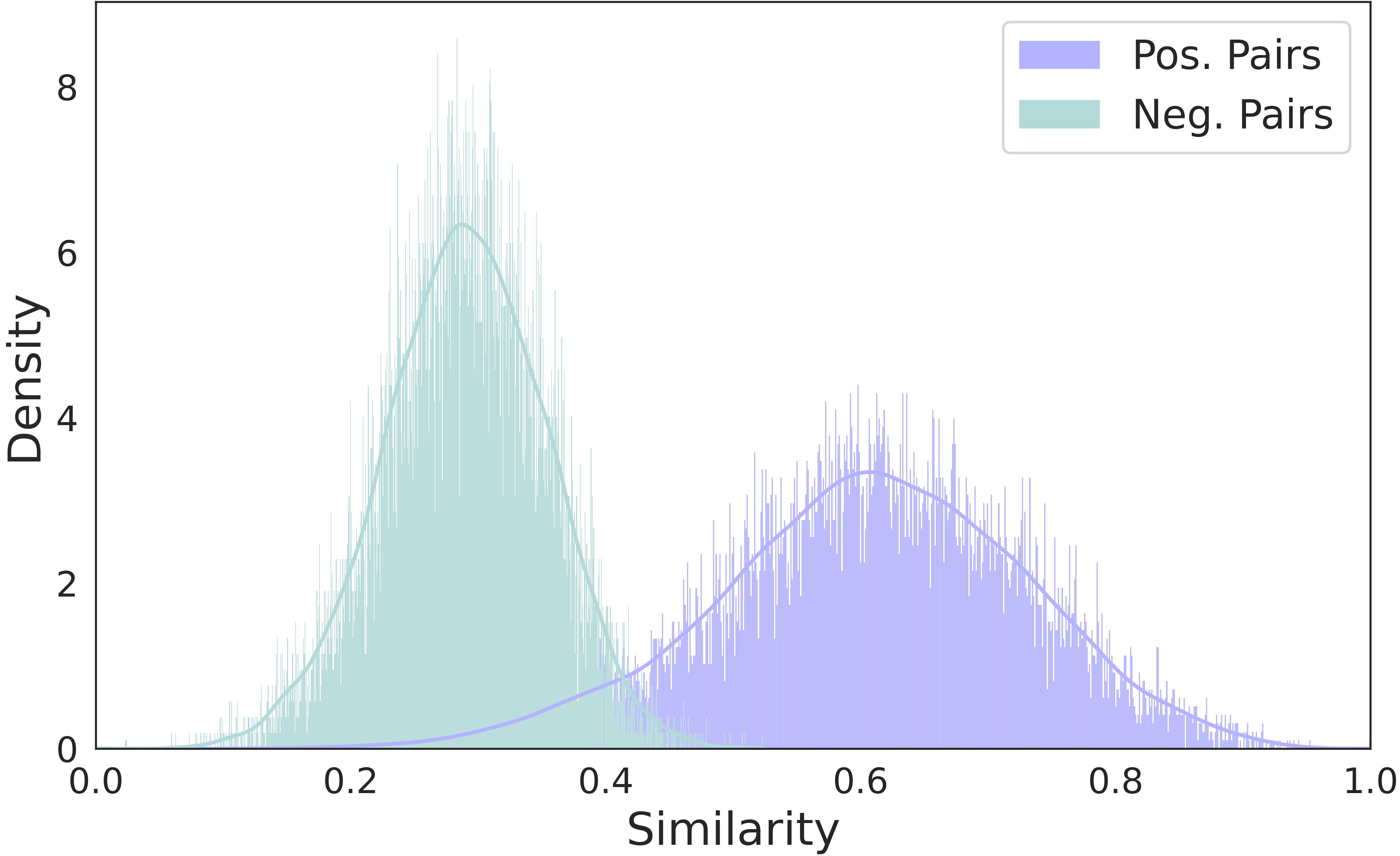}\label{fig: negfull_sims_norm}}\hspace{0cm}
\subfloat[\small TCR Normal]{\includegraphics[width=0.19\textwidth]{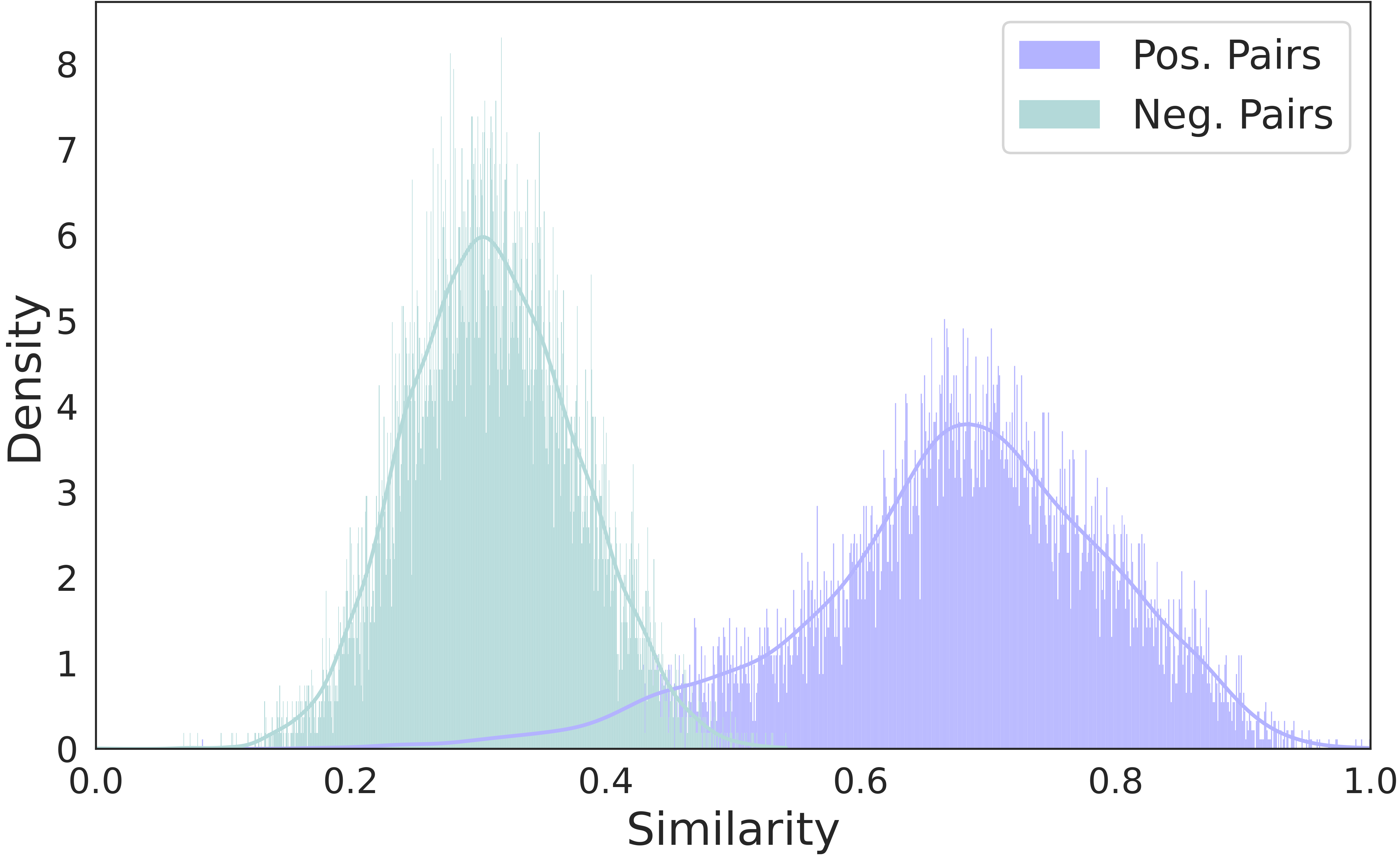}\label{fig: tcr_sims_norm}}\hspace{0cm}
\subfloat[\small NEAT Normal]{\includegraphics[width=0.19\textwidth]{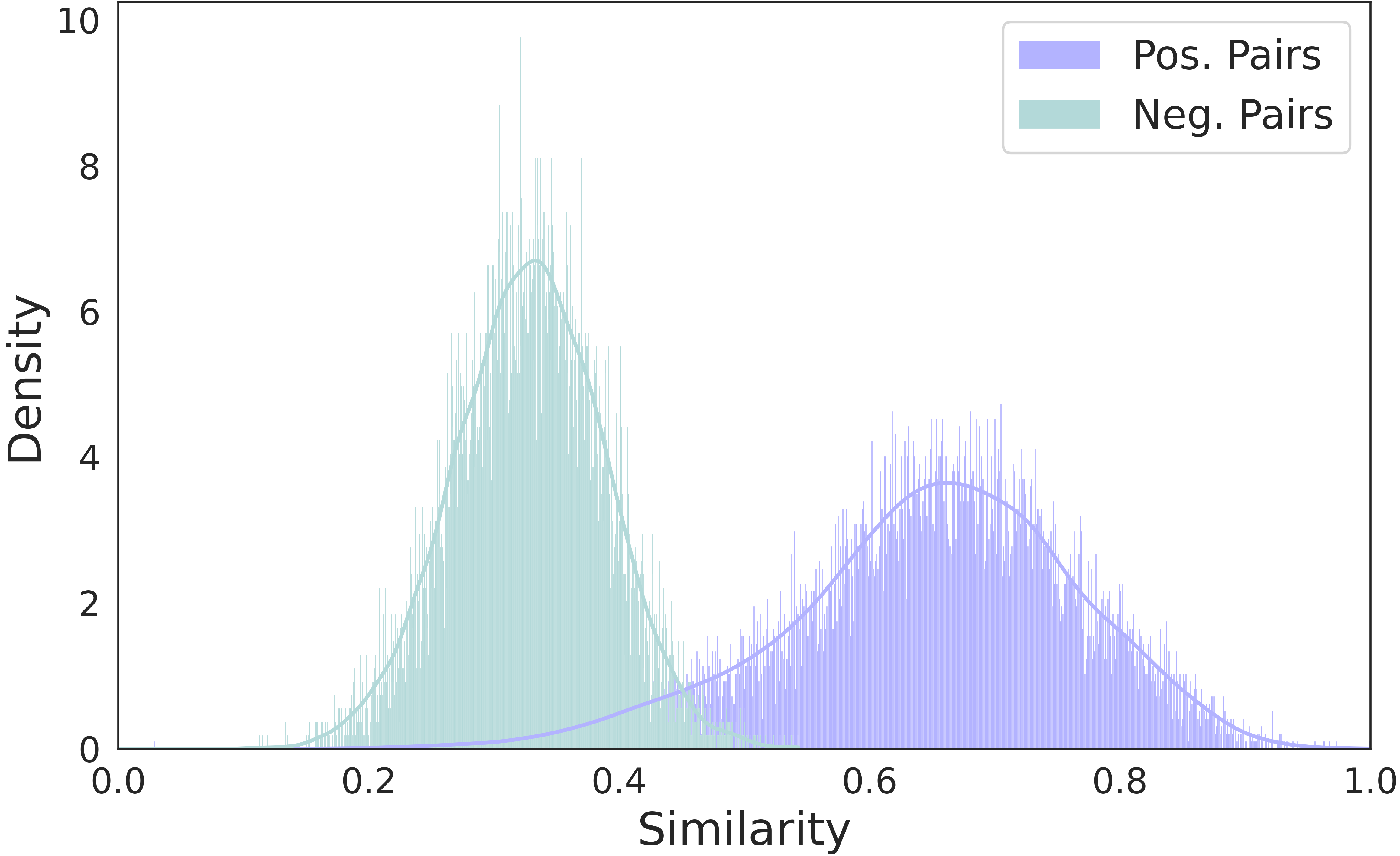}\label{fig: neat_sims_norm}}\\
\vspace{-0.2cm}
\subfloat[\small CLIP NC]{\includegraphics[width=0.19\textwidth]{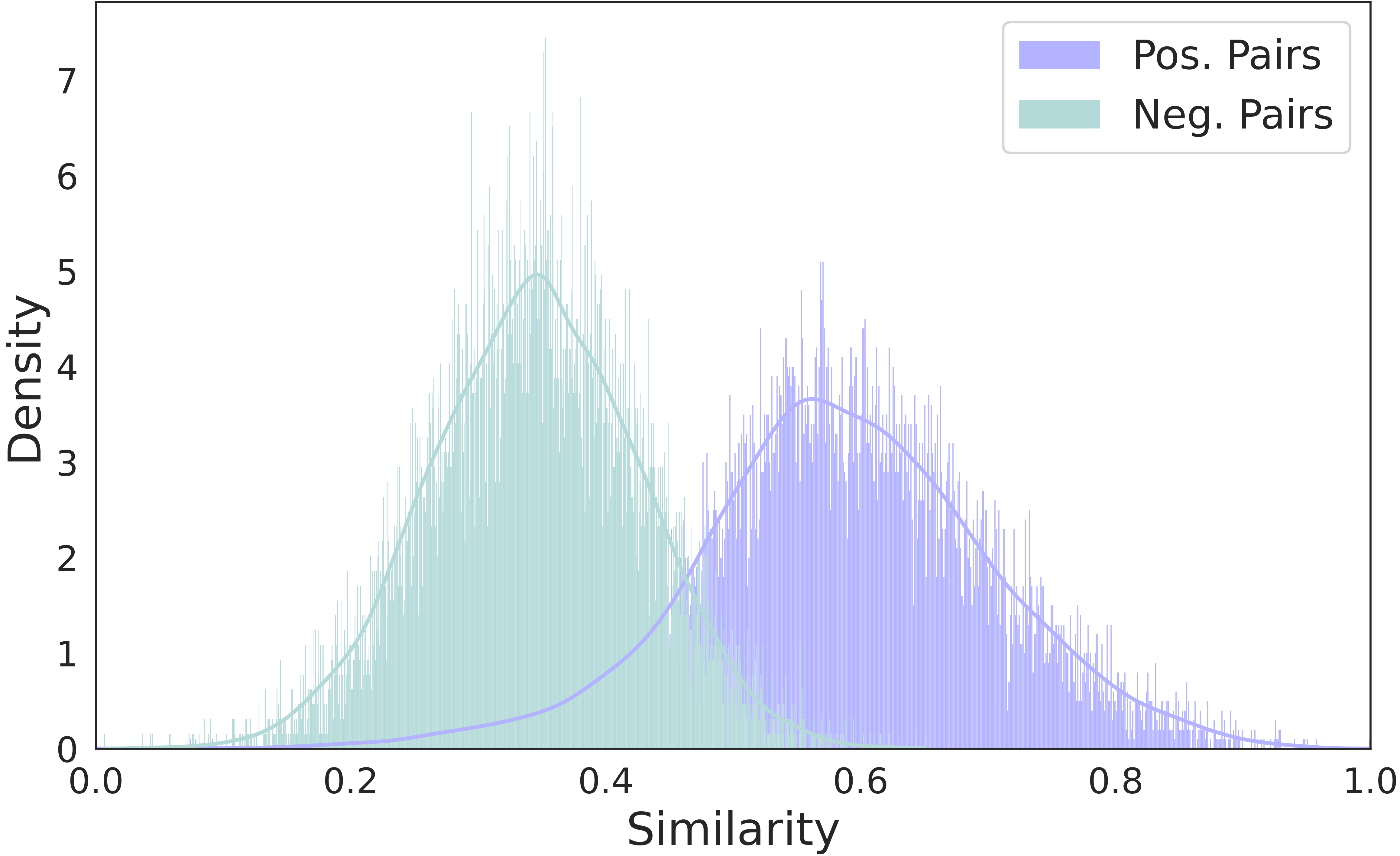}\label{fig: clip_sims_NC}}\hspace{0cm}
\subfloat[\small ConCLIP NC]{\includegraphics[width=0.19\textwidth]{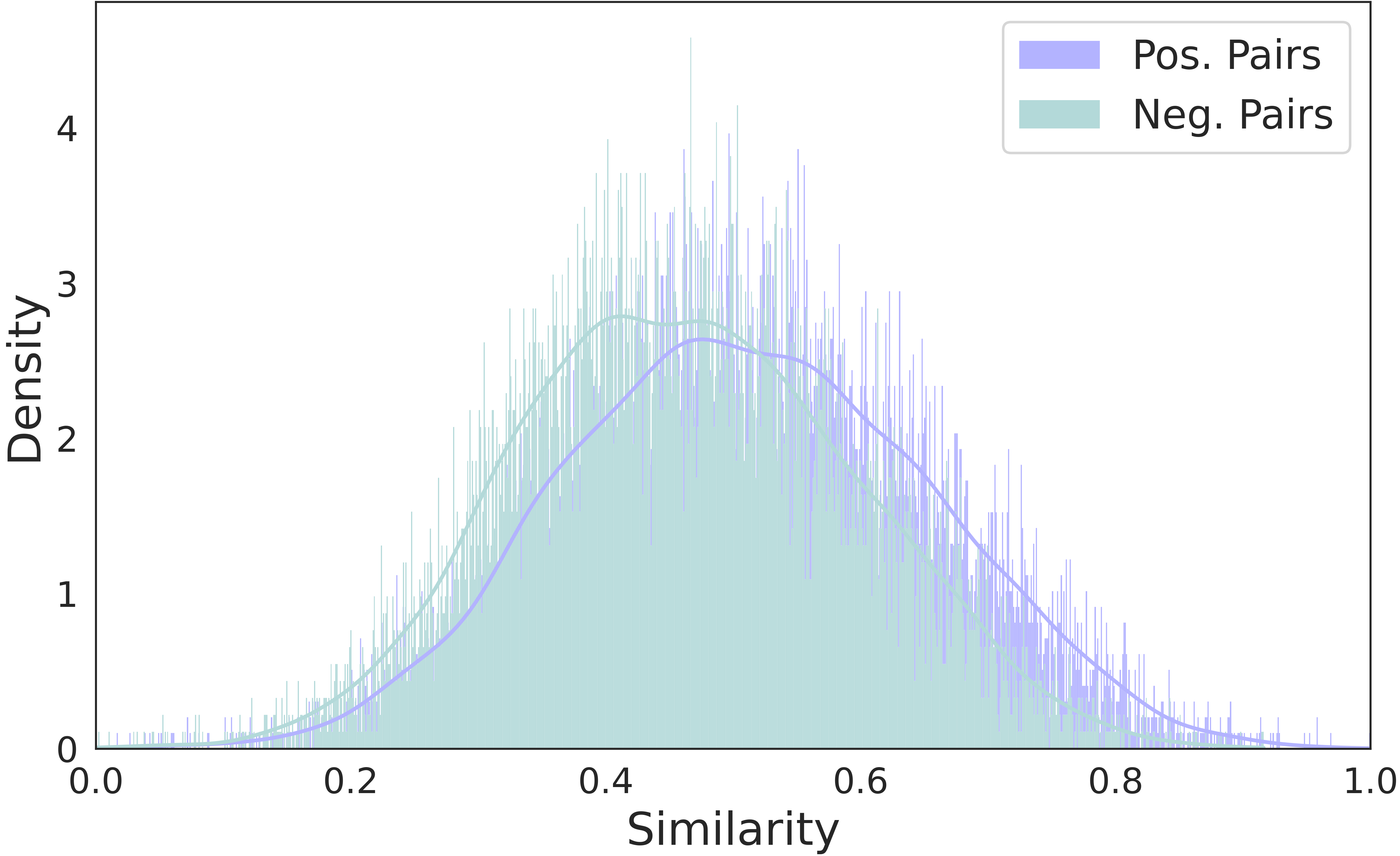}\label{fig: CONclip_sims_NC}}\hspace{0cm}
\subfloat[\small NegFull NC]{\includegraphics[width=0.19\textwidth]{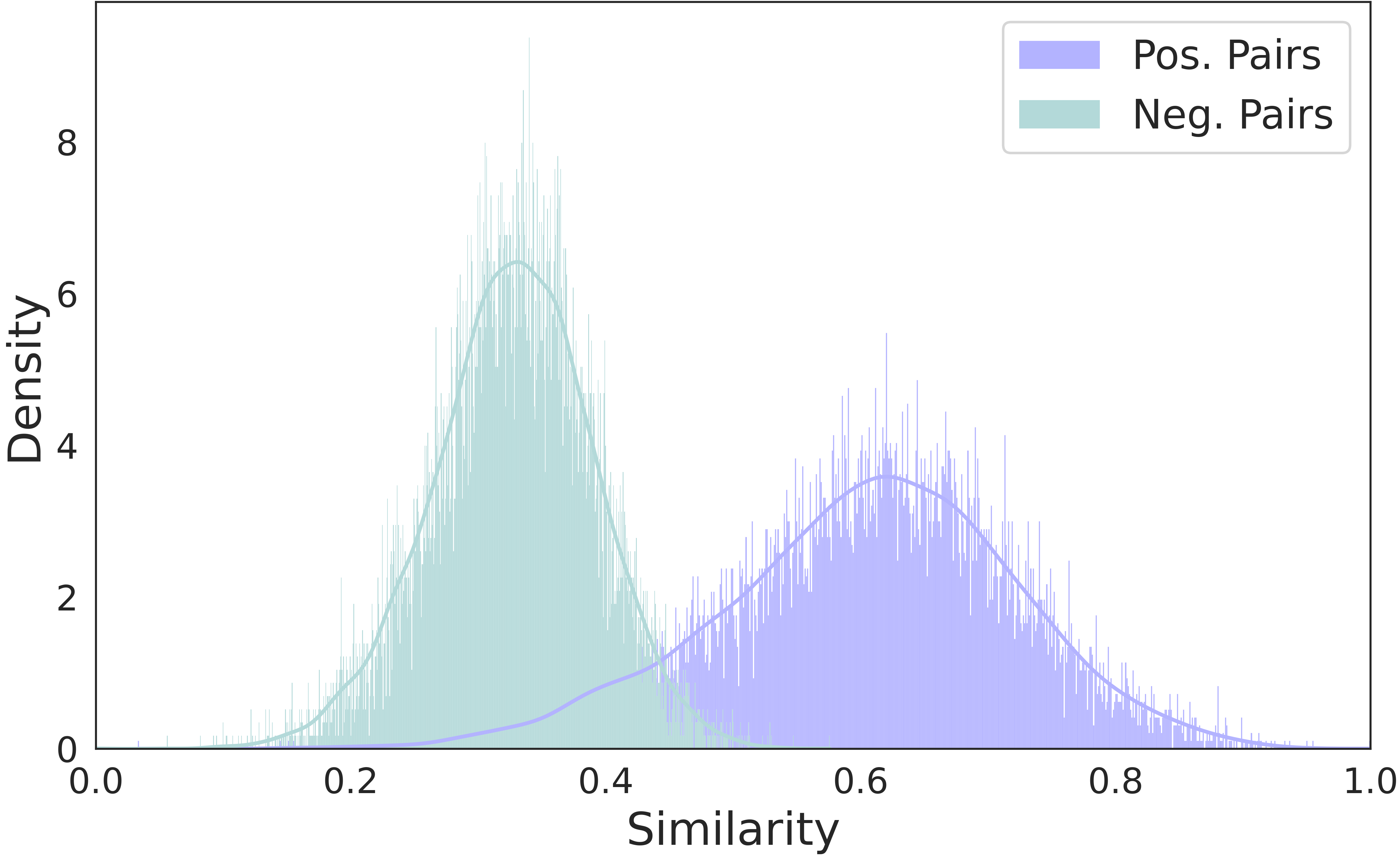}\label{fig: negfull_sims_NC}}\hspace{0cm}
\subfloat[\small TCR NC]{\includegraphics[width=0.19\textwidth]{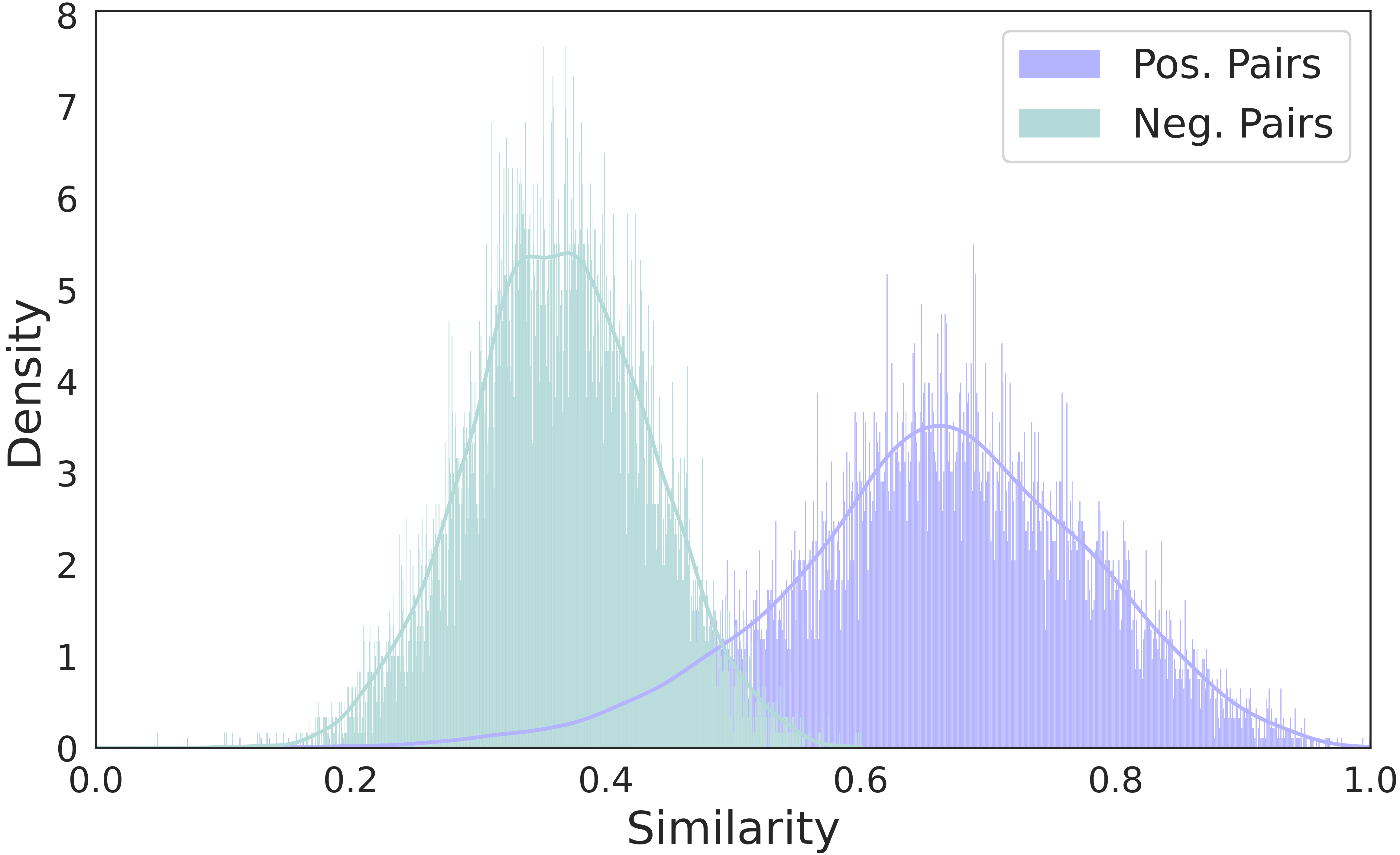}\label{fig: tcr_sims_NC}}\hspace{0cm}
\subfloat[\small NEAT NC]{\includegraphics[width=0.19\textwidth]{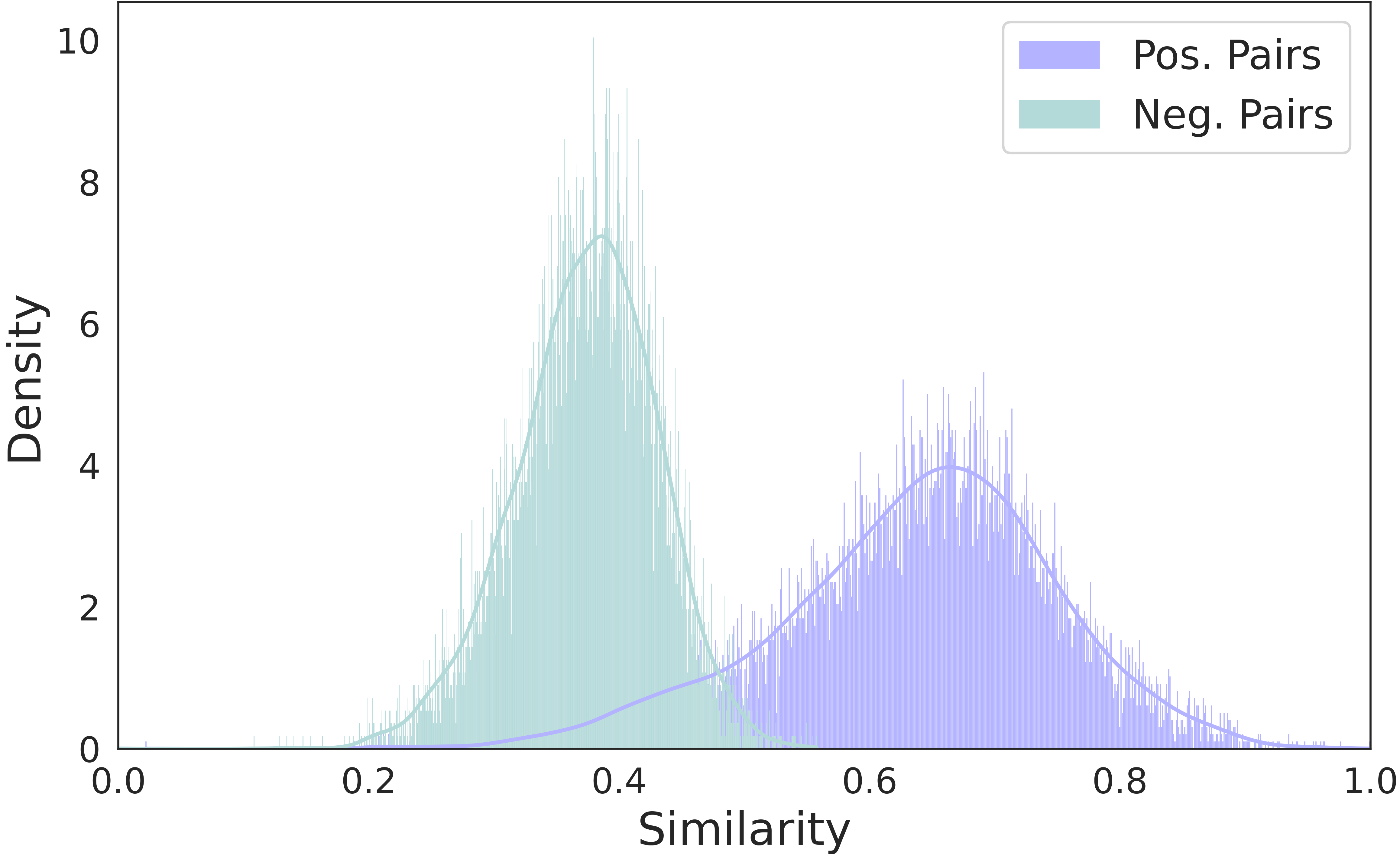}\label{fig: neat_sims_NC}}\\
\vspace{-0.2cm}
\subfloat[\small CLIP RNC]{\includegraphics[width=0.19\textwidth]{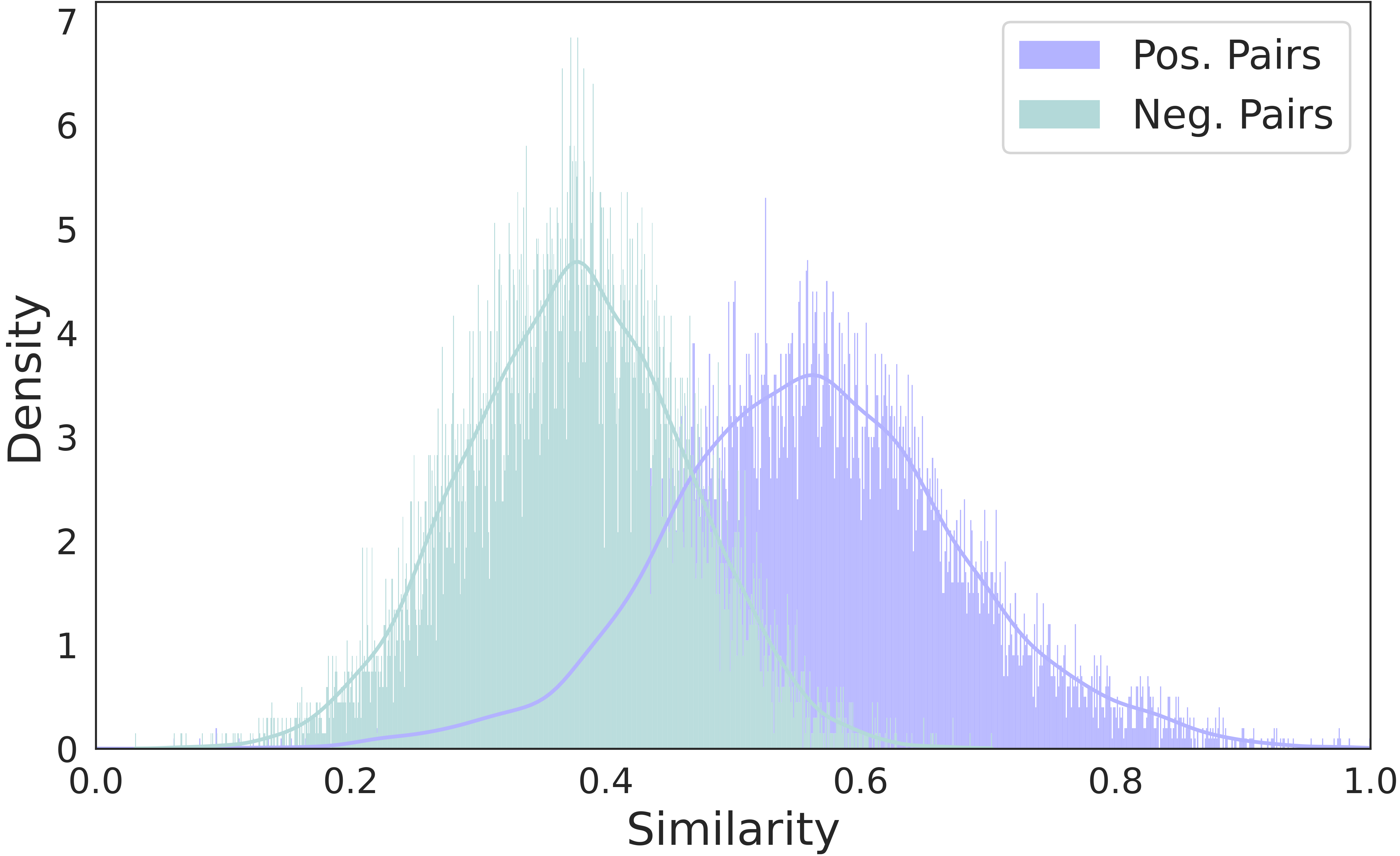}\label{fig: clip_sims_RNC}}\hspace{0cm}
\subfloat[\small ConCLIP RNC]{\includegraphics[width=0.19\textwidth]{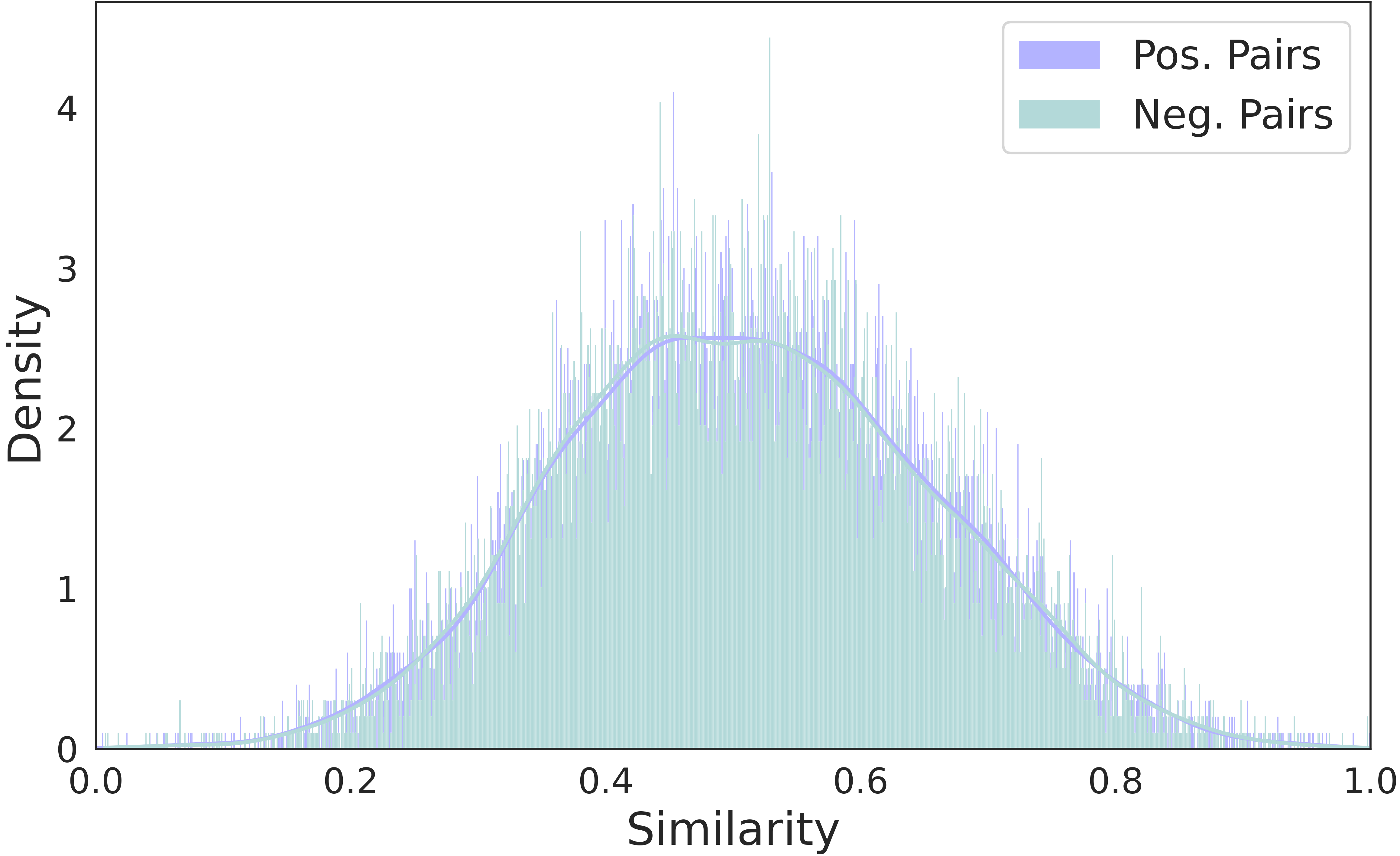}\label{fig: conclip_sims_RNC}}\hspace{0cm}
\subfloat[\small NegFull RNC]{\includegraphics[width=0.19\textwidth]{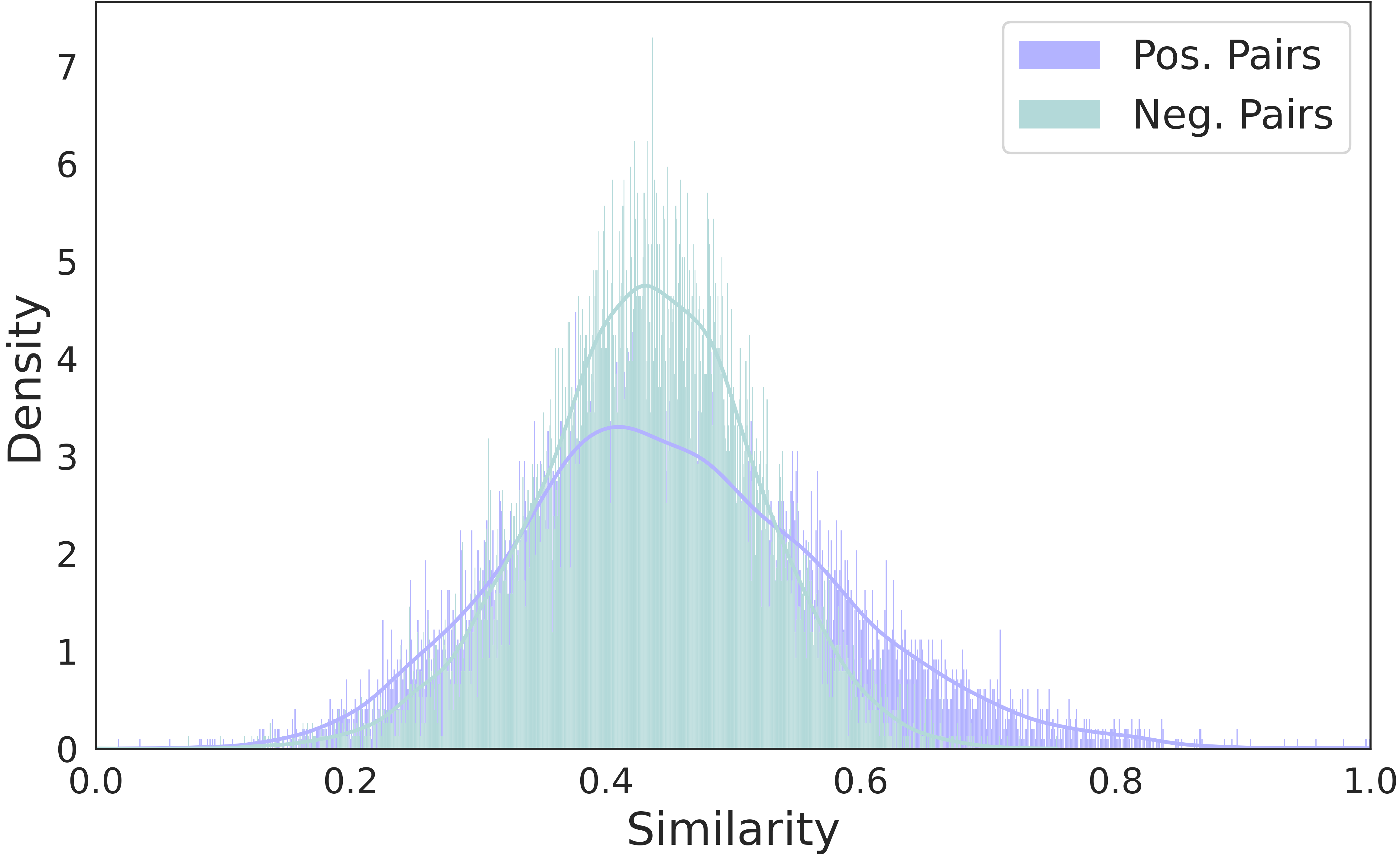}\label{fig: negfull_sims_RNC}}\hspace{0cm}
\subfloat[\small TCR RNC]{\includegraphics[width=0.19\textwidth]{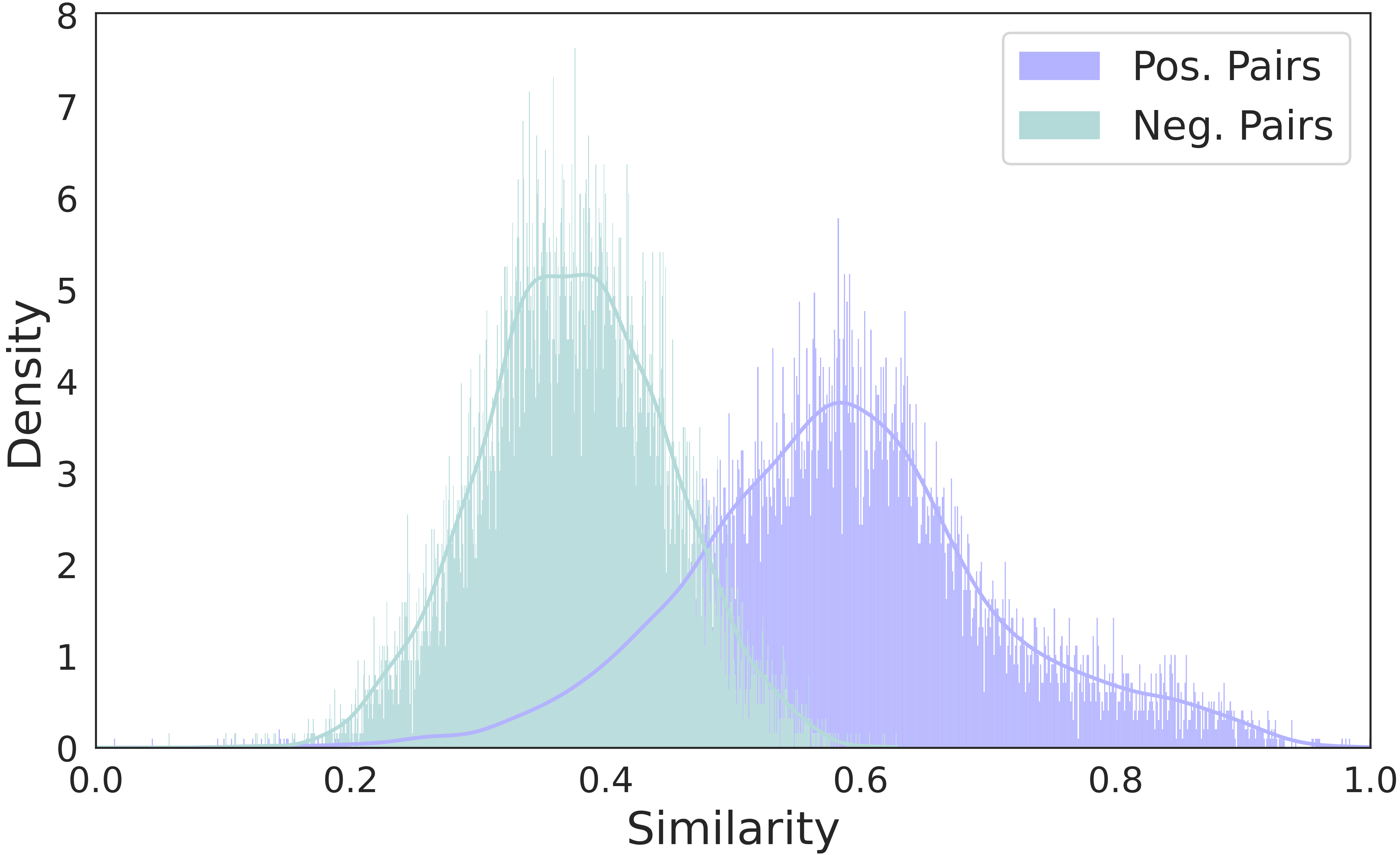}\label{fig: tcr_sims_RNC}}\hspace{0cm}
\subfloat[\small NEAT RNC]{\includegraphics[width=0.19\textwidth]{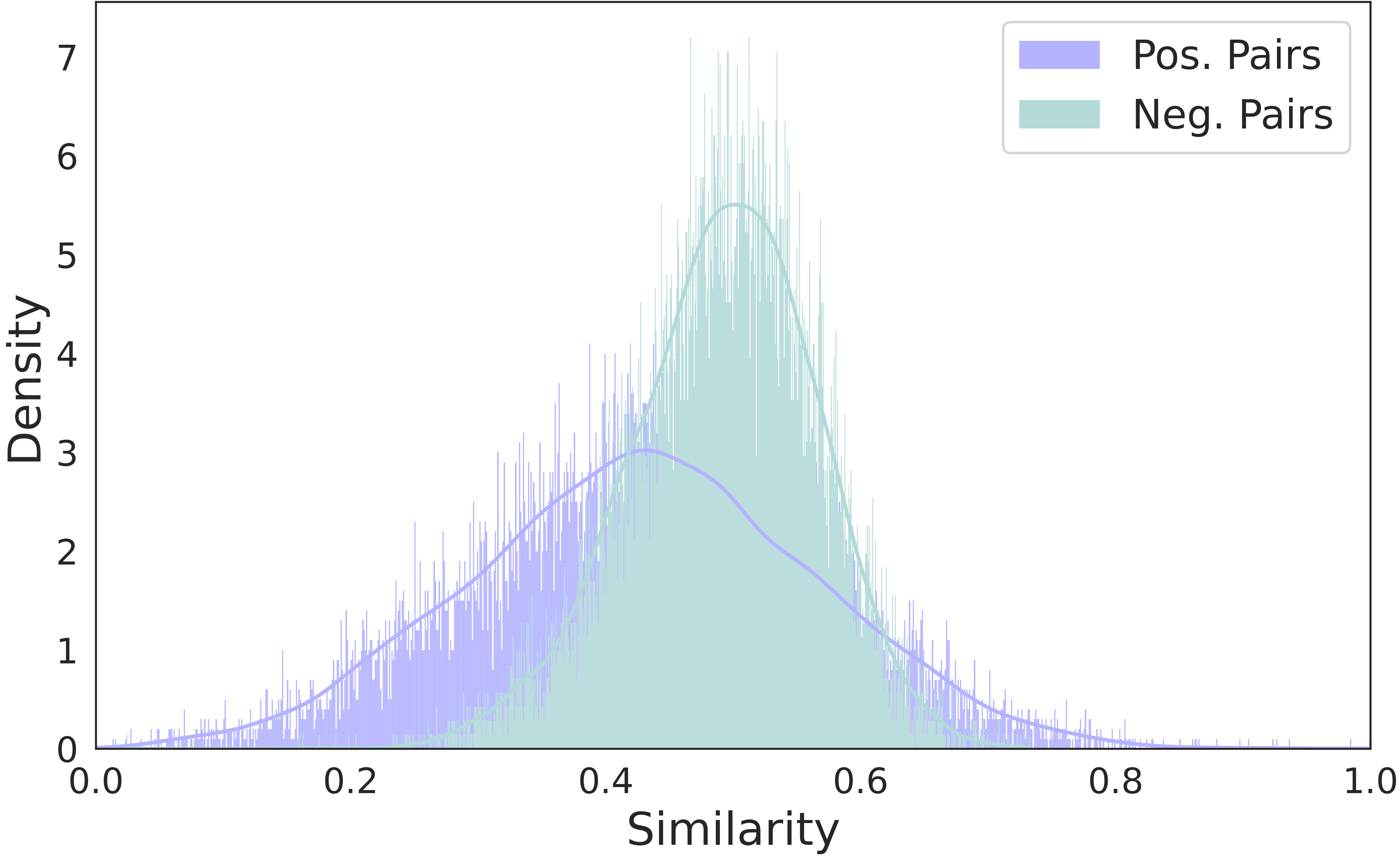}\label{fig: neat_sims_RNC}}
\caption{Comparison of discrimination capability across different negation types. This figure shows the similarity distributions of positive and negative pairs (mean) from VLMs. Image-text similarity distributions for normal, negation-conditioned (NC), and reversed negation-conditioned text (RNC) on the CIFAR10 test set are shown in the top, middle, and bottom rows, respectively. Similarities are normalized to the 0-1 range for clearer presentation.}
\label{fig: visual_sims}
\end{figure*}

\begin{table}[t]
  \centering
    \setlength{\aboverulesep}{-0.58pt}
\setlength{\belowrulesep}{0pt}
\setlength{\extrarowheight}{0pt}
    \caption{Ablation study of the proposed training objects.}
  \renewcommand\arraystretch{1}
  \resizebox{0.95\columnwidth}{!}{
    \begin{tabular}{lll|ccc}
    \toprule
    $\mathcal{L}_{ent}$  & $\mathcal{L}_{sr}$  & $\mathcal{L}_{tri}$  & I2T R@1 & T2I R@1 & MCQ  \\
    \midrule
          &       &       & 45.30 & 24.96 & 39.25 \\
    \checkmark     &       &       & 48.06 & 29.20 & 28.69 \\
          & \checkmark     & \checkmark     & 7.12  & 4.67  & 42.78 \\
    \checkmark     &       & \checkmark     & 49.74 & 28.95 & 27.26 \\
    \checkmark     & \checkmark     &       & 44.96 & \textbf{30.05} & 45.91 \\
    \checkmark     & \checkmark     & \checkmark     & \textbf{48.98} & 30.00 & \textbf{48.41} \\
    \bottomrule
    \end{tabular}%
}

      \label{tab: ablation}%
\end{table}%

\subsubsection{Parameter Analysis}
We next investigate the effect of the parameter $\lambda$ by plotting the online Recall@1 scores with incremental values on MSR-VTT. As shown in Fig. \ref{fig: ablation_lambda}, we observe that: 1) when using smaller $\lambda$ values, \eg, $1$ and $2$, the second term in Eq.\eqref{eq: textual debiasing} would dominate the optimization objective, leading to sub-optimal model performance. 2) Our method can achieve stable performance in a relatively larger range, \ie. $5 \sim 20$, and shows no substantial performance variation even when using extremely large $\lambda$ values. 

\subsubsection{Temperature Sensitivity} As TTA approaches are usually sensitive to temperature parameters, we further carry out experiments to investigate the influence of $\tau_1$ and $\tau_2$. The comparison results are shown in Fig. \ref{fig: ablation_tau}. The figure shows that the adaptation process exhibits different sensitivities to $\tau_1$ and $\tau_2$. Specifically, the adapted model achieves optimal performance with smaller $\tau_1$ values but with larger $\tau_2$ values. Recall that our NEAT enhances negation understanding by reducing the dual-concept shifts, where $\tau_1$ and $\tau_2$ control the distribution shift in consistent and irrelevant semantics, respectively. In Eq.\eqref{eq: refined prediction}, sufficient contrastive targets enable us to utilize sharper distributions (smaller temperature) to enhance the discrimination of the model. However, in Eq.\eqref{eq: contrastive object}, we only sample one hardest contrastive candidate to avoid undesirable clustering. Thus, we need smoother distributions (larger temperature) to prevent overfitting to negative samples.

\subsection{Visualization and Analysis}

\subsubsection{Embedding Similarity Analysis} To visually investigate the performance of our NEAT against negation, we illustrate pairwise similarity distributions of image-text embeddings from different VLMs. Specifically, we compared our NEAT (adapted on unlabeled MS-COCO Retrieval-Neg) with CLIP, ConCLIP, NegFull-finetuned CLIP, and TCR (adapted on unlabeled MS-COCO Retrieval-Neg) in Fig. \ref{fig: visual_sims}. For each image in the test set of CIFAR10, we contrast it with the normal text, \ie, ``a photo of the [CLASS]'', the negation-conditioned text, \ie, ``a photo of the [CLASS] but not of the [CLASS$^{\prime}$]'', and the reversed negation-conditioned text, \ie, ``a photo of the [CLASS$^{\prime}$] but not of the [CLASS]'', where [CLASS] is the true label and [CLASS$^{\prime}$] is a false one. As shown in Figs. \ref{fig: clip_sims_norm}-\ref{fig: neat_sims_norm}, all VLMs could separate the positive and negative pairs apart enough. Although some methods are designed for negation understanding, \eg, ConCLIP and NEAT, their ability to handle affirmative statements remains intact. When tasked with negation-conditioned understanding, Figs. \ref{fig: clip_sims_NC}-\ref{fig: neat_sims_NC} demonstrate that some VLMs exhibit significant overlap between positive and negative distributions, eg, CLIP and ConCLIP. In contrast, our NEAT could discriminate the true and false negative pairs successfully. While for the more challenging reversed negation-conditioned text, Figs. \ref{fig: clip_sims_RNC}-\ref{fig: neat_sims_RNC} show that CLIP and TCR wrongly provide higher similarity scores for positive (the most irrelevant) pairs. Although some negation-enhanced models, \eg, ConCLIP and NegFull-finetuned CLIP, reduce similarity scores for positive pairs, they still maintain distributions comparable to those of negative pairs. Differently, our NEAT is the only approach that properly produces lower similarity scores for positives than negatives, which is consistent with the findings in \Cref{tab: image_classification_results}.

\begin{figure}[tbp]
\centering
\subfloat[\small MSR-VTT MCQ-Neg Task]{\includegraphics[width=\columnwidth]{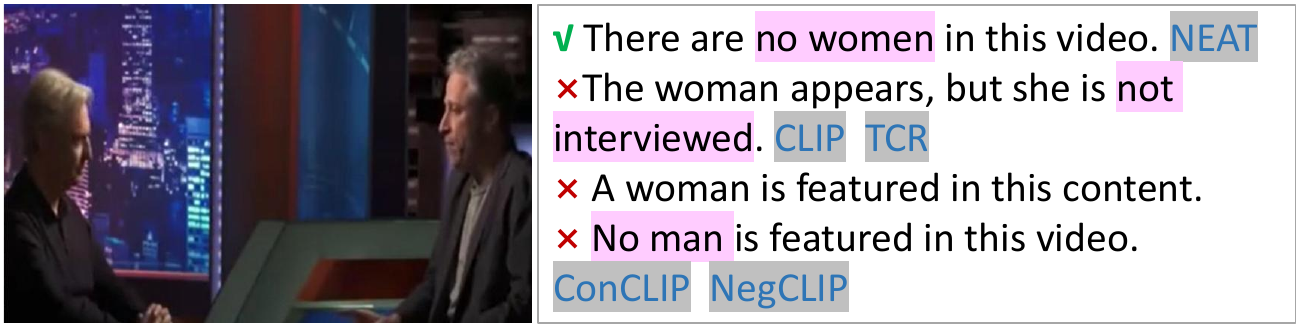}\label{fig: mcq_case}}\\
\vspace{-0.3cm}
\subfloat[\small MSR-VTT Retrieval-Neg Task]{\includegraphics[width=\columnwidth]{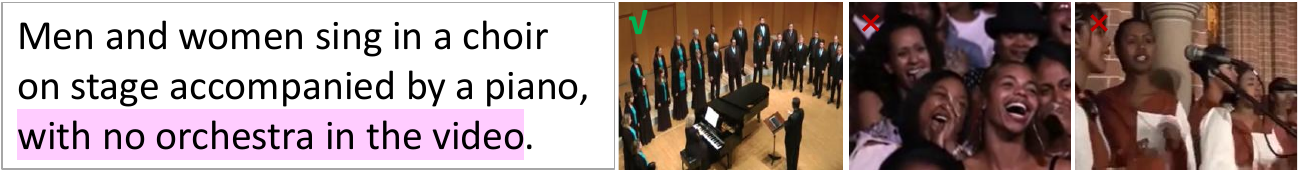}\label{fig: retrieval_case}}
\caption{Case studies of our NEAT in understanding negation. (a) Multiple choice questions with negated captions for a given video. (b) Top-3 retrieved videos for a given negated query. The negation parts are highlighted.}
\label{fig: case_study}
\end{figure}

\subsubsection{Negation Understanding Examples}
To visually illustrate the negation comprehension ability of our NEAT, we show some retrieved videos using negated queries and multiple choice questions with negated captions from MSR-VTT in Fig. \ref{fig: case_study}, where TTA methods (TCR and NEAT) are adapted on unlabeled MSR-VTT Retrieval-Neg data. As shown in the MCQ results in Fig. \ref{fig: mcq_case}, most methods frequently misinterpret negation in either objects (\eg, ConCLIP and NegCLIP) or actions (\eg, CLIP and TCR). From the retrieved results in Fig. \ref{fig: retrieval_case}, one could see that our NEAT is not misled by negation statements and successfully retrieves relevant samples. Overall, the above cases show that our NEAT can rapidly improve performance during test time.

\section{Conclusion}
In this paper, we study enhancing the negation understanding ability of VLMs through test-time adaptation. The key idea is to address the dual-concept shifts problem between affirmation and negation distributions. Specifically, we propose NEAT, a novel method that reduces distribution shift in consistent semantics while eliminating false distributional consistency in unrelated semantics. By adjusting only the lightweight normalization layers, our method could efficiently adapt VLMs to negation contexts during inference. Extensive experiments across multiple benchmarks spanning images, videos, and medical scenarios verify the effectiveness and generalization of our method. In future work, we plan to explore more negation understanding scenarios, \eg, remote sensing image retrieval with negation-conditioned queries, and tasks, \eg, visual question answering, broadening NEAT to handle these corresponding challenges.

\section*{Acknowledgments}
This work was supported in part by the Major Key Project of PCL under Grant PCL2025AS10 and PCL2024A06, and in part by the Shenzhen Science and Technology Program under Grant RCJC20231211085918010.


%

{\small
\bibliographystyle{ieee_fullname}
\bibliography{egbib}
}



\vfill

\end{document}